\pdfoutput=1

\documentclass[11pt]{article}

\usepackage[final]{acl}

\usepackage{times}
\usepackage{latexsym}

\usepackage[T1]{fontenc}

\usepackage[utf8]{inputenc}

\usepackage{microtype}

\usepackage{inconsolata}

\usepackage{graphicx}
\usepackage{amsmath}
\usepackage{subfigure}
\usepackage{enumitem}
\usepackage{amssymb}
\usepackage{booktabs}
\usepackage{multirow}
\usepackage{graphicx}
\usepackage{appendix}
\usepackage{bm}
\usepackage{graphicx}
\usepackage{subcaption}
\usepackage{xspace}
\usepackage{authblk}
\usepackage{hyperref} 

\newcommand{\name}{ADORE\xspace}

\title{Efficient Sparse Attention needs Adaptive Token Release}

\author{
  \textbf{Chaoran Zhang}$^{1}$,
  \textbf{Lixin Zou}$^{1}$\thanks{Corresponding author.},
  \textbf{Dan Luo}$^{2}$,
\textbf{Min Tang}$^{3}$,\\
  \textbf{Xiangyang Luo}$^{4\,\ast}$,
  \textbf{Zihao Li}$^{1}$,
  \textbf{Chenliang Li}$^{1}$\\
  \textsuperscript{1}Wuhan University,
  \textsuperscript{2}Lehigh University,
  \textsuperscript{3}Monash University,\\
  \textsuperscript{4}State Key Lab of Mathematical Engineering and Advanced Computing, China \\
  \{chaoran.zhang, zoulixin, zihao.li, cllee\}@whu.edu.cn, dal417@lehigh.edu, \\ 
  min.tang@monash.edu.cn, xiangyangluo@126.com
}

\begin{document}
\maketitle
\begin{abstract}
In recent years, Large Language Models (LLMs) have demonstrated remarkable capabilities across a wide array of text-centric tasks. However, their `large' scale introduces significant computational and storage challenges, particularly in managing the key-value states of the transformer, which limits their wider applicability.
Therefore, we propose to adaptively release resources from caches and rebuild the necessary key-value states. 
Particularly, we accomplish this by a lightweight controller module to approximate an ideal top-$K$ sparse attention. 
This module retains the tokens with the highest top-$K$ attention weights and simultaneously rebuilds the discarded but necessary tokens, which may become essential for future decoding.
Comprehensive experiments in natural language generation and modeling reveal that our method is not only competitive with full attention in terms of performance but also achieves a significant throughput improvement of up to \textbf{221.8}\%.
The code for replication is available on the \url{https://github.com/WHUIR/ADORE}.
\end{abstract}

\section{Introduction}
\label{sec:introduction}
After breaking through the cognitive barriers, large language models~(LLMs) are now widely used in many text-rich areas, such as voice assistants~\cite{vu2024gptvoicetasker}, search engines~\cite{10.1145/3589334.3645474}, and recommendation systems~\cite{10.1145/3539618.3591778, 10.1145/3543507.3583418}. 
These successes are a testament to the philosophy of scaling up parameters to boost performance, i.e., the scaling law~\cite{kaplan2020scaling}. 
However, in situations demanding rapid or extensive text modeling, the vast size of the model significantly escalates the computational and storage requirements for the key-value~(KV) states of self-attention, which, in turn, limits its throughput~\cite{liu2023scissorhands, strati2024d}.
For example, when using a model with 7 billion parameters, caching the KV states for 1,000 tokens results in a memory requirement that exceeds twice the size of the model parameters, consequently increasing time costs in attention cost and memory swapping.

\begin{figure}[t]
\centerline{\includegraphics[width=0.5\textwidth]
{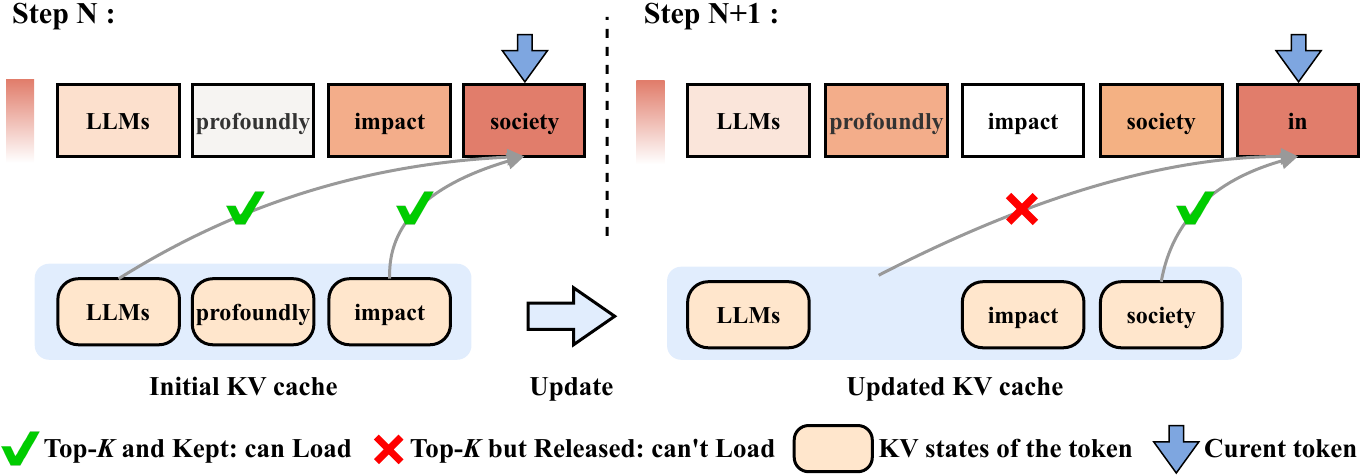}}
\caption{An illustration of the conflict of releasing key-value (KV) states in advance during the inference. Consider a cache size of 3. At step N, the KV states associated with the word `profoundly' are released from the cache. Consequently, in the subsequent step N+1, the `profoundly' state is absent from the cache, despite having a higher attention score for `in'.}
\vspace{-3mm}
\label{fig:introduction}
\end{figure}

Recent efforts address this issue from two perspectives: \textbf{1)} hardware optimization, analogous to `increasing income'; \textbf{2)} refining algorithms, similar to `reducing expenditure'. 
The former approach typically optimizes performance by scheduling tasks across multiple GPUs~\cite{borzunov2022petals} or by implementing hierarchical unloading using the CPU and disk~\cite{aminabadi2022deepspeed, sheng2023flexgen}. 
These techniques, though efficient, require additional hardware and, if not carefully scheduled, can lead to increased communication latency. This, in turn, may potentially degrade the overall user experience~\cite{rasley2020deepspeed, yang2023predictive}.
The latter strategy enhances efficiency by limiting the caching size of key-value states, such as sparsely attending to its immediate neighbors~\cite{zaheer2020big, beltagy2020longformer} or compressing prompts~\cite{li-etal-2023-compressing, pan2024llmlingua}. 
Though efficient, it can often lead to a drop in performance.
Besides, some methods instantiate the sparse attention by masking attention \textit{after} the attention weights have been calculated~\cite{rao2021dynamicvit, li-etal-2023-narrowbert}. Thus, they fail to enhance inference speed and reduce memory usage.  

\vspace{-1mm}
Among existing methods, the dynamic top-$K$ attention~\cite{goyal2020power, 10.1145/3604915.3608779, liang-etal-2023-dynamic}, 
which maintains sparse attention by selecting the highest attention contributions, demonstrates performance comparable to, or even better than, full attention models. 
Such an intuitive solution has only been well explored in encoders where the model can access the entire context simultaneously. 
However, in decoder-based LLMs, the required context shifts dynamically, necessitating multiple calculations of top-$K$ attention throughout the decoding process.
This unique characteristic further complicates the challenges: 
\textbf{1)} Premature and erroneous releasing of unnecessary KV states may result in inaccuracies of top-$K$ attention calculation. However, accurately determining the top-$K$ attention requires considering all KV states of past tokens, thereby conflicting with the goal of reducing costs.
\textbf{2)} Tokens released earlier might be required for top-$K$ attention in future decoding due to long-term dependencies in the text, as illustrated in Figure~\ref{fig:introduction}. Consequently, missing these necessary tokens can result in inaccurate calculations of sparse attention for subsequent tokens.

\vspace{-1mm}
To this end, we introduce \name, \textbf{AD}aptive t\textbf{O}ken \textbf{RE}lease, which maintains a constant cache size by accurately releasing useless past key-value (KV) states and efficiently reconstructing vital past KV states that were previously released.
\name introduces a lightweight controller module that adaptively releases tokens with the lowest predicted attention contribution for the current token from the KV cache. This ensures a fixed KV cache overhead, even when processing numerous tokens. 
Additionally, \name rebuilds the KV state for tokens that are likely to contribute higher attention scores but have been previously released. 
This rebuild mechanism counters the issue when a released token is essential for future decoding.
Moreover, \name can seamlessly integrate into LLM inference, showing impressive results with only minor fine-tuning and training needed for the lightweight controller module. 
Extensive experiments on multiple benchmark datasets reveal that \name achieves up to a 221.8\% improvement in throughput compared to full attention models while maintaining nearly identical text quality.

\section{Methodology}
\label{sec:framework}
\begin{figure*}[h]\centerline{\includegraphics[width=0.9\textwidth]{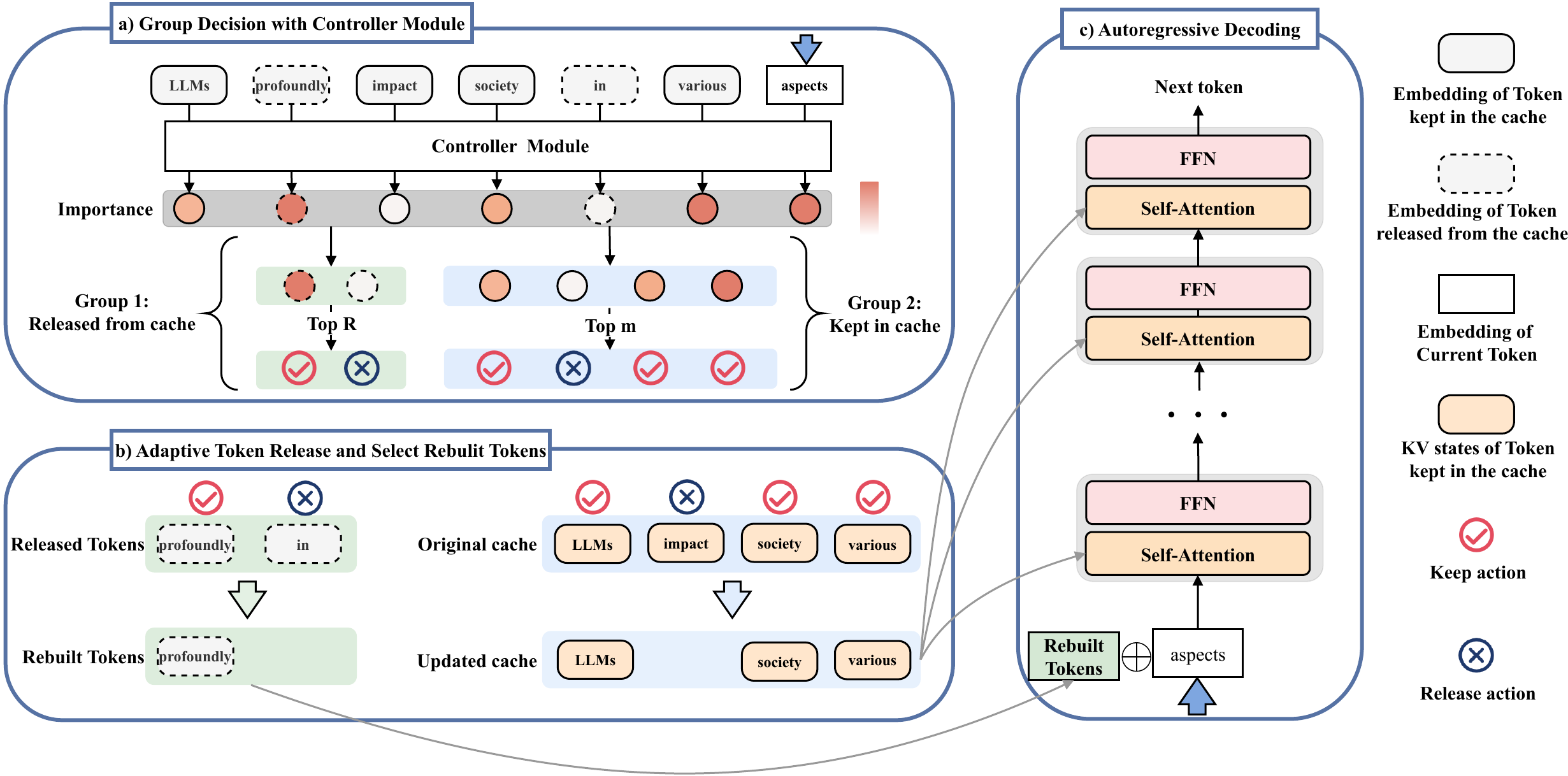}}
\vspace{-3mm}
\caption{The controller module calculates the importance of all input and generated tokens for the current token. The Key-Value (KV) cache maintains the states of $m$ tokens with the highest importance. For tokens that were previously released from the cache, those with the top-$R$ highest importance are concurrently modeled alongside the current token.
}
\label{fig:overall}
\end{figure*}

This section first establishes the framework for efficient sparse attention, followed by initially exploring the adaptive token release in Section~\ref{Adaptive Token Release with Controller Module}. 
Subsequently, we rebuild the KV states of important tokens, approximating the ideal dynamic sparse attention in Section~\ref{Rebuild token's State}. 
Finally, we propose an optimized matrix slicing algorithm to accelerate the implementation of our method in Section~\ref{Batch Expansion}.
An overview of our method is illustrated in Figure~\ref{fig:overall}.

\subsection{Efficient Sparse Transformer} 
\label{Problem Formulation}
Let \( T_n = \{t_{1}, \ldots, t_{s}, t_{s+1}, \ldots, t_{n}\} \) be a set of word tokens, where \( \{t_{1}, \ldots, t_{s}\} \) represent user input tokens, and \( \{t_{s+1}, \ldots, t_{n}\} \) are tokens generated by a transformer-based model, such as GPT-Neo~\cite{gpt-neo} and Llama~\cite{touvron2023llama}. 
When generating the next token $t_{n+1}$, the current token $t_n$ serves as the query input. The $t_n$'s key-value states are based on the following scaled dot-product attention as 

\begin{small}
\begin{eqnarray}\small
\label{equ:attention}
    \bm{a}_{n,l} = {\rm softmax}\left(\frac{\bm{q}_{n,l} \times \left(\bm{K}^{n}_{l}\right)^\top}{\sqrt{d}}\right) \times \bm{V}^{n}_{l},
\end{eqnarray}
\end{small}

\noindent where $\bm{a}_{n,l}\in \mathbb{R}^{d}$ denotes the hidden state at the $l^{th}$ layer of the transformer. 
It undergoes a non-linear transformation process to become the key and value states associated with the token $t$, 
$\bm{q}_{n,l}$ denotes the query vector derived from $t_n$ at the $l^{th}$ layer. 
The terms $\bm{K}^{n}_{l}\in \mathbb{R}^{(n) \times d}$ and $\bm{V}^{n}_{l}\in \mathbb{R}^{(n) \times d}$ represent the key and value states from the current token set $T_{n}$ at the same layer. 
These states are retained in the GPU memory to minimize redundant computations.
The generation of the token $t_{n+1}$ is accomplished through a multi-classification approach, utilizing the hidden state $\bm{a}_{n, L}\in \mathbb{R}^{d}$ from the last layer. 

For an efficient sparse transformer, we selectively cache the most relevant KV states, aiming to reduce computational demands while maintaining or even enhancing the model's performance in generating subsequent tokens as 

\begin{small}
\begin{eqnarray}\nonumber\small
\label{eqn:target}
\bm{a}'_{n,l} = {\rm softmax}\left(\frac{\bm{q}_{n,l} \times \left(\bm{K}^{n}_{m+1, l}\right)^\top}{\sqrt{d}}\right) \times \bm{V}^{n}_{m+1,l}.
\end{eqnarray}
\end{small}

\noindent Here, $\bm{a}'_{n,l}$ approximates the $\bm{a}_{n,l}$ using the $\bm{K}^{n}_{m+1, l}\in \mathbb{R}^{(m+1)\times d}$, $\bm{V}^{n}_{(m+1),l}\in \mathbb{R}^{(m+1)\times d}$, which correspond to selecting $m$ rows from $\bm{K}^{n-1}_{l}$ and $\bm{V}^{n-1}_{l}$ and concatenating them with $\bm{k}_{n,l}$ and $\bm{v}_{n,l}$ respectively, with the condition that $m << n-1$. $\bm{k}_{n,l}$, $\bm{v}_{n,l}$ denote the key and value vector derived from $t_n$ at the $l^{th}$ layer.
It implies only a significantly smaller $\bm{K}^{n}_{m+1, l}$ and $\bm{V}^{n}_{m+1,l}$ are retained in GPU for rapid inference and save memory. 

From a performance standpoint, achieving the ideal sparsity involves computing the full attention weight $\bm{w}_{n} = \bm{q}_{n,l} \times \left(\bm{K}^{n}_{l}\right)^\top \in \mathbb{R}^{n}$ and then selecting the top-$m$ query-key product weights.
Then, these weights serve as indices for slicing $\bm{V}^{n}_{l}$. 
While this method is optimal in performance, it does not confer any computational or memory savings as the process of computing full attention weights for all query-key pairs and then selecting the top weights is computationally intensive.

\subsection{Adaptive Token Release} 
\label{Adaptive Token Release with Controller Module}
The adaptive token release is to create efficient scheduling of the key-value states within the GPU memory. 
The main idea is to use a lightweight controller module as an alternative to computing full weight for slicing the full key-value states. 
To be both efficient and effective, we have implemented several design strategies:
\begin{itemize}[leftmargin=0.1pt]
\item \textbf{Refine the model with top-$K$ attention.} Compared to the full attention, Top-$K$ could mitigate the impact of excluding partial KV states once the pertinent top-$K$ KV states are included within the $m$ cached KV states, which is consistent with the target defined in~Equ (\ref{eqn:target}).
Therefore, we initially fine-tune the LLMs with top-$K$ attention, which utilizes only the highest top-$K$ attention weights via masking the weights out of the top-$K$.
Remarkably, this approach yields performance that is on par with full attention models~\cite{liu2022dynast}. 
To be efficient, the cache size $m$ is slightly larger than $K$. As $m$ decreases, the complexity of the scheduling process increases correspondingly.
\item \textbf{Adopt a uniform scheduling policy for the retention or exclusion of KV states across various layers.} Constructing a layer-specific scheduling strategy would necessitate additional time to model each layer's input. 
Moreover, the initial layer is more pivotal for integrating value states; as we delve deeper into the layers, the hidden states become increasingly homogeneous~\cite{wu2023ppt}.
Additionally, it is observed that different layers often focus on a similar set of top-$K$ attentions.
The effectiveness of the uniform scheduling policy is elaborated in Appendix~\ref{Analysis of Dynamic Sparse Attention}.
\item \textbf{Update the cached KV states by appending the latest KV state and selectively release an older one.} 
An intuitive idea is to store the KV states in the motherboard's memory as backup. However, due to bandwidth limitations between the GPU and motherboard, moving KV states in and out proves to be extremely slow, at times even slower than recalculating the KV states~\cite{aminabadi2022deepspeed}. Consequently, when updating the cached KV states, we simply append the most recently computed KV states while removing a nonsignificant older one, thereby maintaining a constant size for the cache.
\end{itemize}

Adhering to above strategies, we develop a controller module that utilizes the lightweight and efficient GRU~\cite{dey2017gate} for scheduling the cached KV states. 
Specifically, during the generation of token $t_{n+1}$, we establish the probability of whether caching the KV state of token $t_{i}$ as:

\begin{small}
\begin{eqnarray}
    \bm{z}_{i} &=& {\rm GRU}(\bm{x}_{i}, \bm{z}_{i-1}) \\ \nonumber
    \sigma_{i} &=& {\rm Sigmoid}({\rm MLP}(\bm{p}_{i}+\bm{z}_{i})) 
\end{eqnarray}
\end{small}

\noindent where $\bm{x}_{i}\in \mathbb{R}^{d}$ represents the token embedding from the LLMs. 
The GRU is a single-layer GRU (an unidirectional model with its effectiveness analyzed in Appendix~\ref{Comparison of unidirectional and bidirectional GRU performance}) that recurrently transforms this token embedding into a context-aware representation $\bm{z}_{i}\in \mathbb{R}^{d'}$. 
The term $\bm{p}_{i}\in\mathbb{R}^{d'}$ denotes the position embedding for the $i^{th}$ token, which signifies the importance of token position in the scheduling model. 
During the update of the KV states, we discard those with the lowest $\sigma_{i}$ values and append the most recent KV states to the cached states. 
To fine-tune its parameters, we construct a dataset by collecting word embeddings of each sequence as input. Then we construct corresponding labels by assigning a value of $1$ to the indices of the top-$K$ tokens that most frequently occur within the top-$K/2$ attention scores across all layers, and a value of $0$ to all others.

\subsection{KV States Rebuild}
\label{Rebuild token's State}
Adaptive token releasing facilitates the selective preservation of the most pertinent tokens, yet previously discarded tokens may become essential for future decoding due to the long-term dependencies in text. To counter this issue, we propose the rebuilding of KV states as a complement. 

This method entails retrieving the top-$R$ tokens with the highest $\sigma_{i}$ values from the set of released tokens.
Let $\bm{X}_{R}\in \mathbb{R}^{R \times d}$ represent the token embedding of selected released tokens. 
We concatenate $\bm{X}_{R}$ with $\bm{x}_{n}$, i.e., the embedding of current token $t_n$, forming the input $\bm{X}_{R+1} \in \mathbb{R}^{(R+1) \times d}$.
After $(l-1)$-layers processing, we can obtain the query states $\bm{Q}^{n}_{R+1,l} \in \mathbb{R}^{(R+1) \times d}$, $\bm{K}^{n}_{m+R+1,l}\in \mathbb{R}^{(m+R+1) \times d}$ and $\bm{V}^{n}_{m+R+1,l}\in \mathbb{R}^{(m+R+1) \times d}$, where $\bm{K}^{n}_{m+R+1,l}$/$\bm{V}^{n}_{m+R+1,l}$ is formulated by concating cached key/value state and rebuild key/value states for the input tokens. 
With its argument, the attention is calculated as

\begin{scriptsize}
\begin{eqnarray}
\nonumber
\bm{A}'_{R+1,l} = {\rm softmax}\left(\frac{\bm{Q}^{n}_{R+1,l} \times \left(\bm{K}^{n}_{m+R+1, l}\right)^\top}{\sqrt{d}}\right) \times \bm{V}^{n}_{m+R+1,l},  
\end{eqnarray}
\end{scriptsize}

\noindent where $\bm{A}'_{R+1,l}$ is the hidden state. To get the corresponding value for the current generating tokens, we get the $\bm{a}'_{n,l}$ by selecting the last row of $\bm{A}'_{R+1,l}$. 
Through the parallel rebuilding of the released KV states, we maximize the utilization of GPU without incurring excessive time overhead.

\subsection{Matrix Slicing as Multiplication}
\label{Batch Expansion}
The scheduling of KV states relies on certain matrix-slicing operators. 
Traditional slicing operators like \texttt{gather} and \texttt{mask-select} can lead to significant time overheads~\cite{kim2022analysis}, particularly when batch operations involve varying slicing indices. 
To circumvent it, we leverage the GPU's rapid matrix multiplication capabilities. 
For instance, to remove the $j^{th}$ row from $\bm{K}^{n}_{m,l}$, we prepare a slicing matrix, $S_j = I_{(1:j-1,j+1:m), :}$, where $I\in \mathbb{R}^{m\times m}$ is the identity matrix and $I_{(1:j-1,j+1:m), :}$ selects all rows of $I$ except the $j^{th}$ row. 
The resulting $\bm{K}^{n}_{m-1,l} = S_j \times \bm{K}^{n}_{m,l}$, with $S_j$ being pre-prepared to save time.

\section{Experiment}
\label{sec:experiment}

\subsection{Experimental Settings}

\begin{table*}[h]
\scriptsize
\centering
\setlength{\tabcolsep}{3.3mm} 
\begin{tabular}{l|ccc|ccc|ccc}
  \bottomrule
Dataset & \multicolumn{3}{c|}{UltraChat} & \multicolumn{3}{c|}{EverythingLM} & \multicolumn{3}{c}{Math}\\
Metric &BLEU&ROUGE&BERT-F&BLEU&ROUGE&BERT-F&BLEU&ROUGE&BERT-F\\
  \hline\hline
  \multirow{1}{*}{Full Attention} & 35.6 & 29.2 & 63.4 & 35.4 & 30.8 & 64.5 & 38.6 & 29.9 & 69.7\\ \hline
  \multirow{1}{*}{Window Attention} & 26.7 & 28.0 & 61.4 & 22.3 & 25.9 & 62.3 & 30.3 & 24.3 & 66.3 \\
  \multirow{1}{*}{Strided Attention} & 28.0 & 24.8 & 57.5 & 20.3 & 22.1 & 58.5 & 33.0 & 26.7 & 66.7 \\
  \multirow{1}{*}{KV Compression} & 21.1 & 23.2 & 56.9 & 18.2 & 22.3 & 55.7 & 32.2 & 24.4 & 66.4 \\
  \multirow{1}{*}{StreamingLLM} & 23.9 & 26.0 & 59.6 & 20.5 & 25.6 & 61.4 & 32.9 & 26.8 & 68.3 \\
  \multirow{1}{*}{H$_{2}$O} & 26.4 & 25.3 & 60.3 & 24.6 & 25.1 & 61.8 & 33.3 & 26.2 & 68.1 \\
  \multirow{1}{*}{H$_{2}$O(Rebuilt)} & 27.6 & 26.9 & 61.5 & 25.2 & 25.6 & 62.1 & 34.7 & 27.1 &  69.6$^{\ast}$ \\
  \multirow{1}{*}{ADORE} & \textbf{36.8$^{\ast}$} & \textbf{28.8} & \textbf{63.5$^{\ast}$} & \textbf{30.4$^{\ast}$} & \textbf{27.7$^{\ast}$} & \textbf{63.1$^{\ast}$} & \textbf{38.8$^{\ast}$} & \textbf{28.9$^{\ast}$} & \textbf{70.5$^{\ast}$} \\
  \toprule
\end{tabular}
\vspace{-3mm}
\caption{Performance comparison of different methods in natural language generation tasks. 
We use Full Attention as the upper limit. The best results are marked \textbf{bold}. 
``$\ast$'' indicates significant improvement over the top-performing sparse attention method, with a $p$-value $<0.01$.}\label{tab:NLG-table}
\end{table*}

\textbf{Dataset.} To evaluate the effectiveness of various sparse attention mechanisms, 
we conduct extensive experiments across three distinct tasks: natural language generation, stream generation, and natural language modeling. 
For the first task, we evaluate on UltraChat~\cite{ding2023enhancing}, EverythingLM\footnote{\url{https://huggingface.co/datasets/totally-not-an-llm/EverythingLM-data}}, and Math~\cite{li2023camel}. 
For the second task, we experiment on StreamEval~\cite{xiao2024efficient} and StreamChat (built upon UltraChat). 
For the last task, we evaluate models on CNN Dailymail~\cite{see-etal-2017-get} and SAMSum~\cite{gliwa-etal-2019-samsum}. 

Specifically, \textbf{UltraChat} is a multi-turn dialogue dataset containing approximately 696,600 training samples and covering diverse topics such as questions about the world and creative writing. 
\textbf{EverythingLM} is an instructional dataset consisting of 1,000 conversations and encompassing a wide array of topics and interactions.
\textbf{Math} dataset is composed of 50,000 problem-solution pairs obtained using GPT-4 across 25 math topics.
\textbf{StreamChat} concatenates every 100 samples from UltraChat and feeds them into the model in a streaming fashion to assess the quality of the generated answers. 
\textbf{StreamEval} is a question-answer dataset, building upon LongEval~\cite{li2023long1}. Specifically, it comprises about 2,000 samples, each with 1,000 lines of textual information and 100 retrieval questions.
\textbf{CNN Dailymail} is a news summarization dataset containing over 300,000 news articles. 
\textbf{SAMSum} is a summarization dataset containing about 16,000 messenger-like conversations with summaries. The details of the datasets are reported in Appendix~\ref{sec:Dataset Statistics}.

\textbf{Baseline.}
We compare our method with the following methods: 
\textbf{1) Full Attention} encompasses all past KV states across every layer, characterized by a time complexity of $O(T^2)$ and linear growth in cache size. 
\textbf{2) Window Attention}~\cite{hassani2023neighborhood} focuses on the nearest tokens for self-attention at each layer, thus ensuring a constant size for the key-value cache.
\textbf{3) Strided Attention}~\cite{child2019sparsetransformer} attends to both the nearest and distant tokens by periodically focusing on one with a fixed interval, thus striking a balance between effectiveness and efficiency. \textbf{4) KV Compression}~\cite{ren-etal-2023-context} incrementally compress the intermediate activation of a specified span of tokens into compact ones.
\textbf{5) StreamingLLM}~\cite{xiao2024efficient} extends Window Attention by adding the first four tokens to the cache, aiming to maintain a normal distribution of attention scores and stable inference settings. 
\textbf{6) H$_{2}$O}~\cite{zhang2024h2o} dynamically evicts unimportant tokens from the cache, which contribute the least to the cumulative attention.
\textbf{7) H$_{2}$O(Rebuilt)} selectively rebuilds the KV states of evicted token based on H$_{2}$O.

\textbf{Experimental Protocols.}
We employ Llama-2 7B~\cite{touvron2023llama} as our backbone for evaluation. It has 32 transformer layers and 4,000 context length. 
For our experiments, we employ the top-$96$ attentions and set the KV cache size $m$ to $192$ with top-$8$ rebuilt tokens.  
We randomly selected 1,000 samples from the benchmark dataset for training purposes. The samples were utilized to develop the sparse top-$K$ backbone model using QLoRA~\cite{dettmers2023qlora}, along with the controller module. The extra data were employed for testing models. The training and inference times for the controller module are detailed in Appendix~\ref{sec:implement-detail} and Appendix~\ref{Comparison of unidirectional and bidirectional GRU performance}, respectively.
To evaluate the quality of the generated text, we use metrics including BLEU, ROUGE, BERT-F~\cite{zhang2019bertscore} and Accuracy. To measure the inference speed of different methods, we use \textbf{Throughput}~\cite{sheng2023flexgen}, which is defined as the number of tokens generated per second. 

\subsection{Natural Language Generation}
\begin{figure}[t]
\centerline{\includegraphics[width=0.35\textwidth]{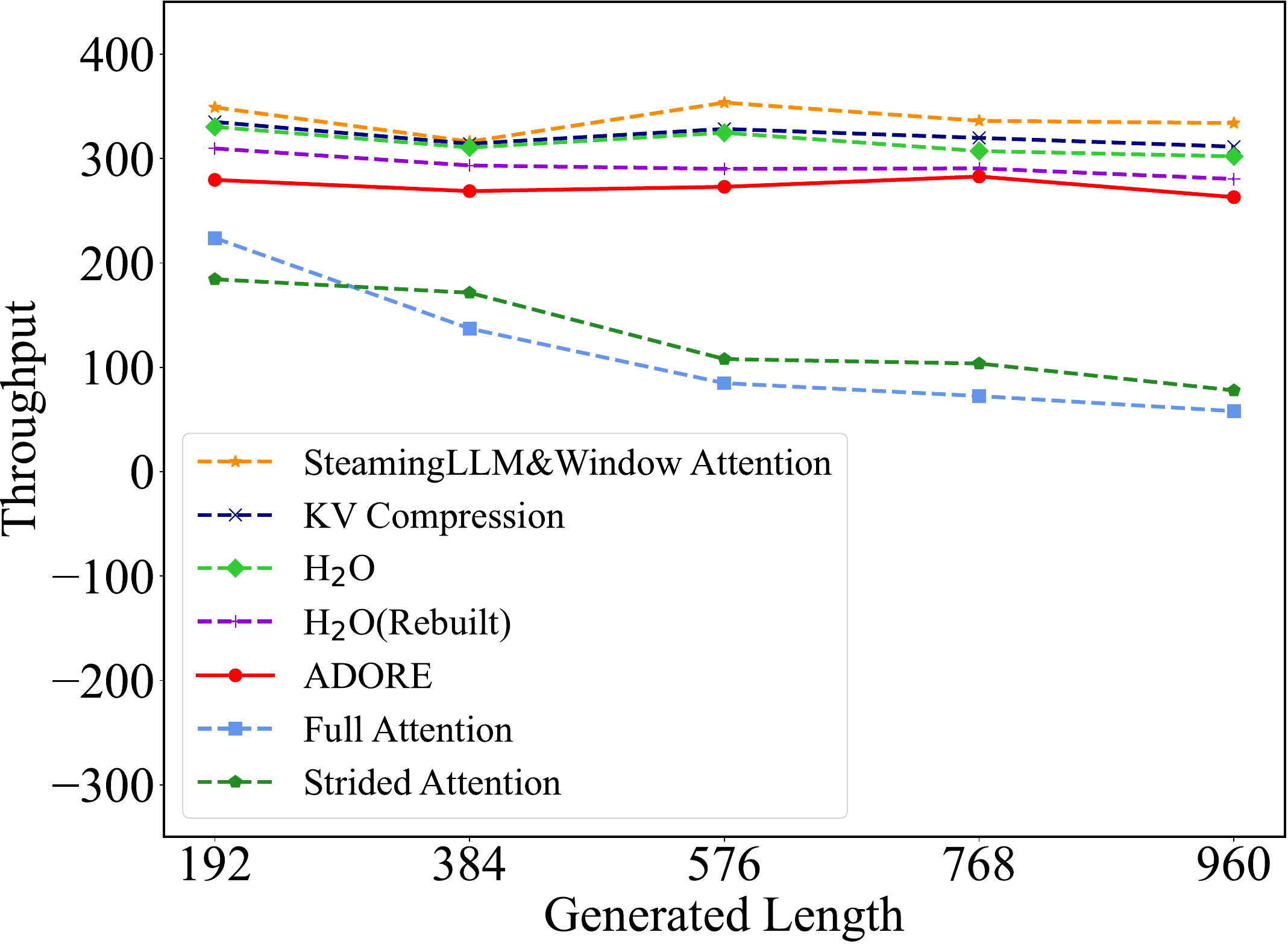}}
\vspace{-3mm}
\caption{Performance comparison in terms of throughput for generating different text lengths.}
\label{fig:thoughtput}
\end{figure}

This subsection evaluates models' performance in natural language generation.
We summarize the quality of generating text on UltraChat, EverythingLM and Math benchmarks in Table~\ref{tab:NLG-table} and the throughput against different sequence lengths in Figure~\ref{fig:thoughtput}. 
From the results reported, we have the following observations: \textbf{1) The proposed \name achieves the best performance, and consistently outperforms all the baselines on all datasets.}
In Table~\ref{tab:NLG-table}, our method shows an improvement over full attention in the UltraChat dataset, with increases of 1.2\% in BLEU scores and 0.1\% in BERT-F scores. On the other hand, Window Attention, Strided Attention, KV Compression, StreamingLLM, H$_{2}$O, and H$_{2}$O(Rebuilt) show reductions of 8.9\%, 7.6\%, 14.5\%, 11.7\%, 9.2\% and 8.0\% in BLEU scores, respectively. A similar trend is also observed in the learning curve illustrated in Appendix~\ref{Analysis of Dynamic Sparse Attention}. \textbf{2) Our proposal performs the best in achieving a high efficiency while maintaining a competitive performance against full attention.}
Specifically, it is evident that our method demonstrates a consistent throughput against various generated text lengths; whereas full attention suffers from a significant drop in throughput as the generated text length increases.
Notably, our method outperforms full attention by 151.4\% and 221.8\% when generating text lengths of 768 and 960, respectively. 
\textbf{3) Existing SOTA methods, i.e., StreamingLLM and H$_{2}$O,  sacrifice inference quality in exchange for improving throughput.} Though StreamingLLM and H$_{2}$O have slightly higher throughput, their performance on natural language generation suffers a lot.

\subsection{Stream Generation}
\label{sec:stream_gen}
\begin{figure}[h]
\centerline{\includegraphics[width=0.35\textwidth]{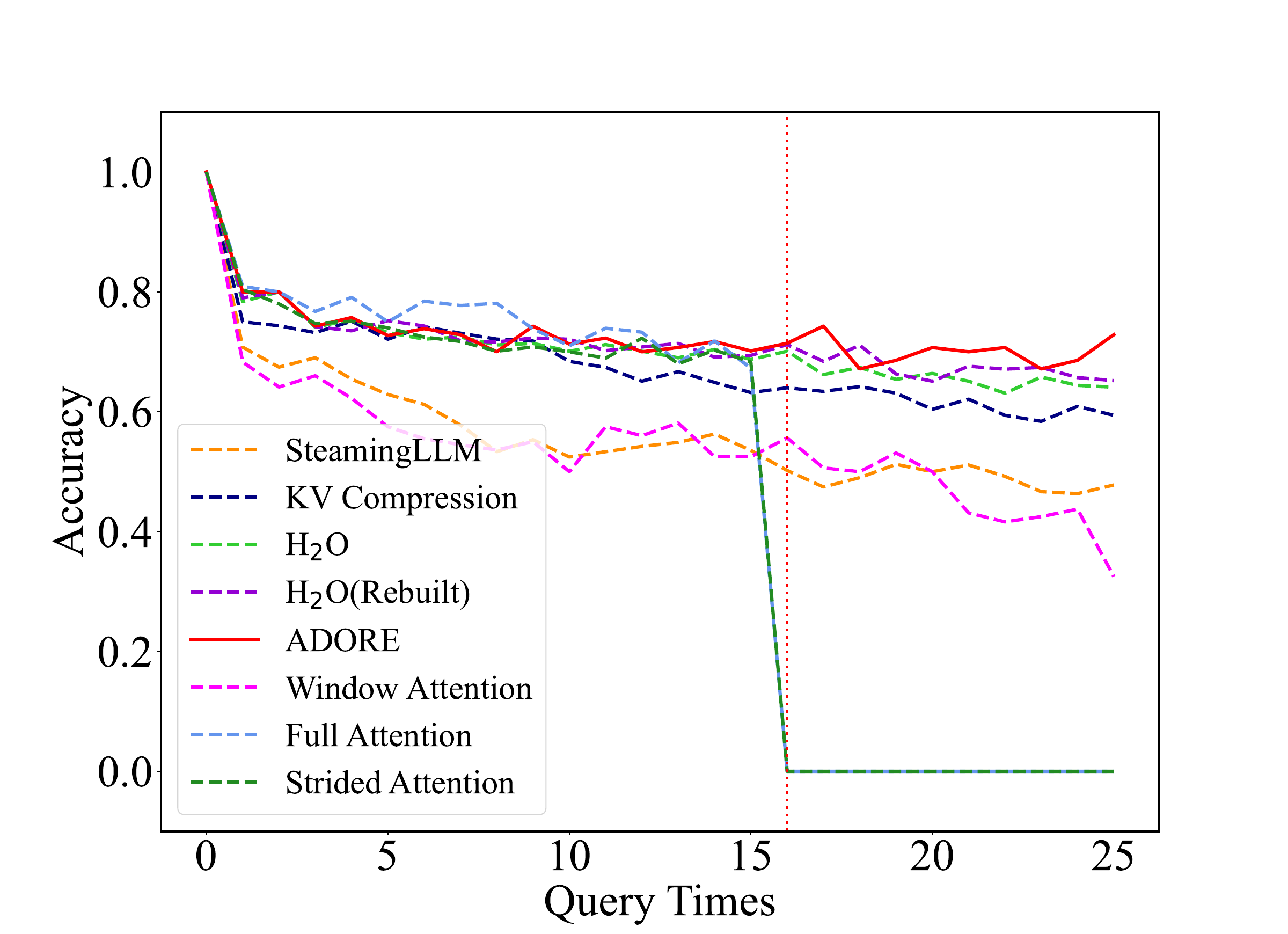}}
\vspace{-3mm}
\caption{Performance comparison on the StreamEval at various query times.}
\label{fig:qa-acc}
\end{figure}

To show the real-world applicability of our proposal, we emulate the performance of the models on infinite streaming dialogue, i.e., StreamChat, and question-answering tasks, i.e., StreamEval.
For StreamChat, we chunk the streaming chat with the size of $4096$ to evaluate the quality of generation against different sequence lengths. 
The experimental results are reported in Table~\ref{tab:stream}. 
For StreamEval, we report the generating accuracy of models' responses after multi-times query in Figure~\ref{fig:qa-acc}. 

\begin{table*}[h]
\scriptsize
\centering
\setlength{\tabcolsep}{0.8mm}
\begin{tabular}{l|ccc|ccc|ccc|ccc}
  \bottomrule
Sequence Length & \multicolumn{3}{c|}{(0, 4096]} & \multicolumn{3}{c|}{(4096, 4096$\times$2]} & \multicolumn{3}{c|}{(4096$\times$2, 4096$\times$3]} & \multicolumn{3}{c}{(4096$\times$3, 4096$\times$4]}\\ 
Metrics &\small{BLEU}&\small{ROUGE}&\small{BERT-F}&\small{BLEU}&\small{ROUGE}&\small{BERT-F}&\small{BLEU}&\small{ROUGE}&\small{BERT-F}&\small{BLEU}&\small{ROUGE}&\small{BERT-F}\\
  \hline\hline
  Full Attention&{42.8}&{43.8}&{70.9}&3.6&4.9&33.1&2.1&2.4&30.1&2.0&2.4&30.0 \\ \hline
  Strided Attention&27.7&30.5&60.9&2.1&3.0&29.7&2.0&2.2&30.0&2.0&2.1&29.6 \\ 
  Window Attention&24.7&28.5&60.6&14.2&19.6&54.3&16.1&18.1&50.1&19.9&20.4&52.1\\
KV Compression&21.2&32.7&61.4&18.3&24.5&59.1&16.4&21.5&56.2&16.9&22.0&57.4\\
StreamingLLM&14.6&36.1&64.8&14.8&29.0&63.2&18.6&28.4&62.4&21.7&27.5&63.5\\
H$_{2}$O&27.4&33.5&65.2&28.3&32.7&64.9&26.2&32.5&64.2&27.3&32.4&65.6\\
H$_{2}$O(Rebuilt)&31.2&35.7&66.0&30.7&34.3&65.8&28.9&33.2&64.7&30.4&34.9&66.7\\
ADORE&\textbf{38.9}&\textbf{38.3}&\textbf{66.4}&\textbf{36.5}&\textbf{39.2}&\textbf{67.7}&\textbf{35.5}&\textbf{37.7}&\textbf{67.1}&\textbf{36.7}&\textbf{39.5}&\textbf{69.1}\\
  \toprule
\end{tabular}
\vspace{-3mm}
\caption{Performance comparison on StreamChat across different lengths. The best results are shown in \textbf{bold}.}\label{tab:stream}
\end{table*}

From the Table~\ref{tab:stream} and Figure~\ref{fig:qa-acc}, we have following observations: 
\textbf{1)} In the table, our method demonstrates a consistent performance across different sequence lengths, which justifies its efficacy in streaming dialogue, especially in length extrapolation and capturing high-importance tokens. 
While full attention exhibits the best performance on the first subset~(length in range $(0, 4096]$), its performance rapidly declines as the streaming sequence length surpasses the pre-training window size, and eventually becomes almost $0$. 
\textbf{2)} In the figure, our method consistently maintains high accuracy, even when the number of queries exceeds 20, which expresses the superiority of our proposed method. 
On the other hand, full attention and strided attention display competitive performance at limited query times. However, they suffer a significant drop in performance due to Out-of-Memory (OOM) issues, which arise as the accumulation of excessive KV states increases with the number of queries.
This observation justifies the necessity of sparse attention.
However, Window Attention and StreamingLLM demonstrate lower accuracy compared to our approach, primarily due to their fixed heuristic policies. KV Compression exhibits suboptimal accuracy due to the information loss in token compression. Benefiting from dynamic eviction strategy, H$_{2}$O and H$_{2}$O (Rebuilt) consistently demonstrate competitive performance compared to our method.

\subsection{Natural Language Modeling}
\label{secsec:Natural Language Modeling}
We evaluate the performance of various methods in natural language modeling on the CNN Dailymail and SAMSum datasets. We report perplexity (ppl.) as the metric to compare the performance of different methods across different sequence length subsets. Similar to Section~\ref{sec:stream_gen}, the length in each subset is in the range of $((i-1)\times1024, i\times1024]$ for $(i=1, 2, \dots)$.

\begin{figure}[htbp]
\centerline{\includegraphics[width=0.5\textwidth]{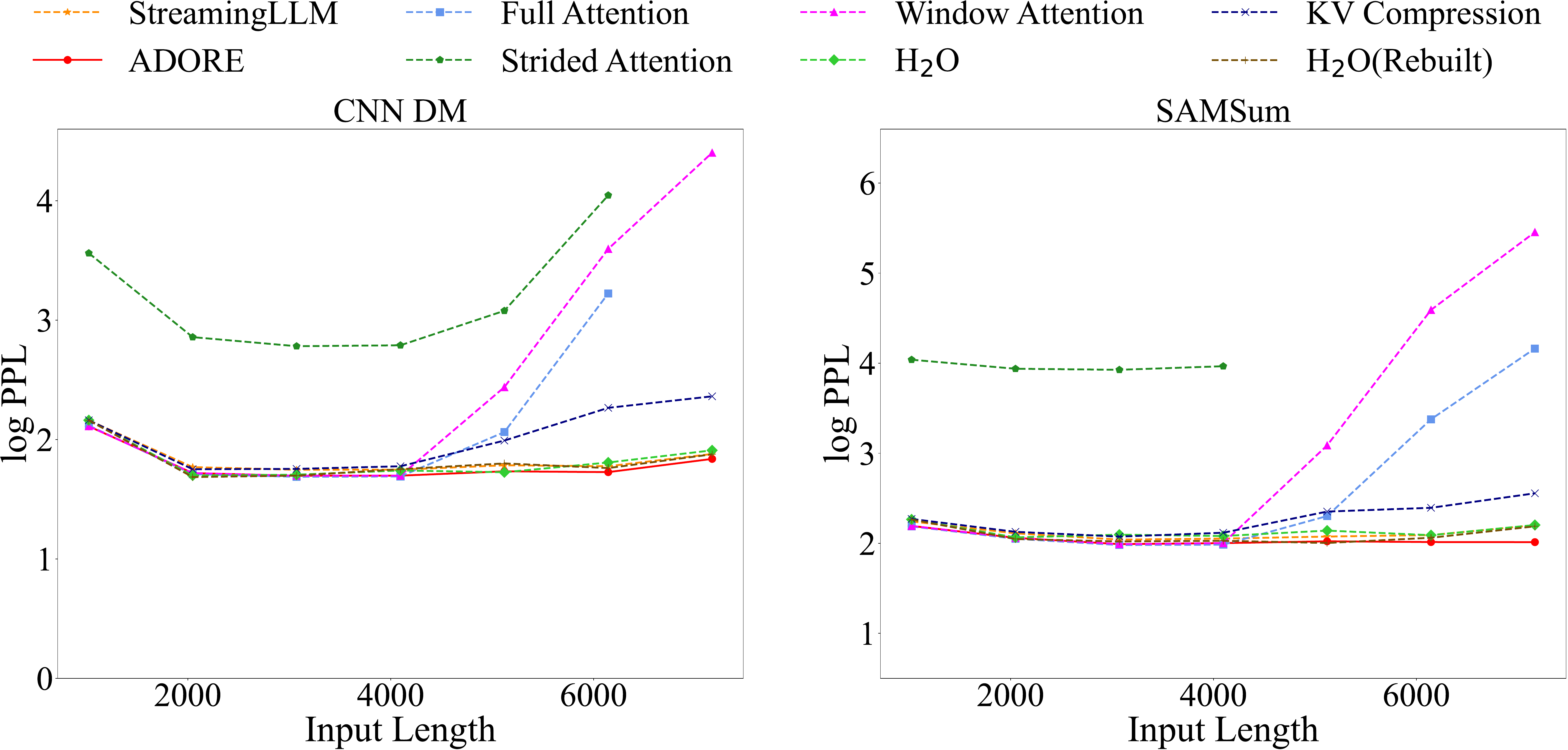}}
\vspace{-3mm}
\caption{Perplexity evaluation on CNN DM and SAMsum across different lengths.}
\label{fig:exp3-ppl}
\end{figure}
Figure~\ref{fig:exp3-ppl} illustrates the logarithm of perplexity for different methods across various modeling intervals. 
It is evident that our method, StreamingLLM, H$_{2}$O, and H$_{2}$O(Rebuilt) consistently maintain the lowest perplexity. They are effective in preserving the original attention distribution with sparse attention. Therefore, they demonstrate superior performance on extrapolating length.
As the sequence length increases, KV Compression compresses more tokens, leading to a gradual increase in perplexity.
Although full attention exhibits the best performance in the shortest input length subset~($[0, 4096]$), its performance quickly becomes worse when the input length surpasses the size of the pretraining window.

\subsection{Ablation Study}
\label{secsec:Ablation Studies}
\subsubsection{Influence of Attention Sparisity}

\begin{figure}[htbp]
\centerline{\includegraphics[width=0.35\textwidth]{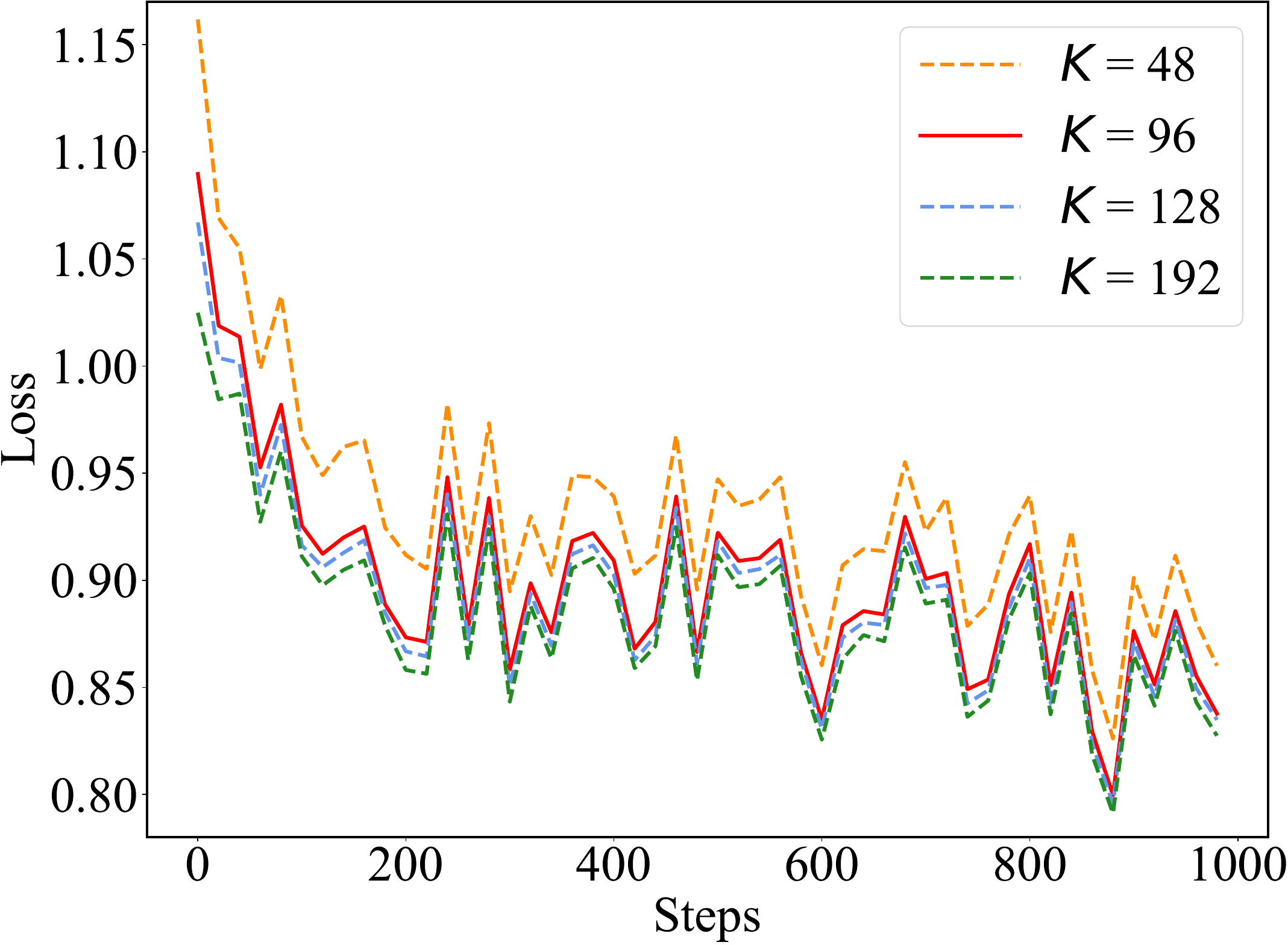}}
\vspace{-3mm}
\caption{Comparison of fine-tuning loss against different values of $K$.}
\label{fig:ablation1}
\end{figure}

We explore \name's performance against different $K$ in adaptive token release. 
In particular, we configure $K$ in the range of \{48, 96, 128, 192\} for fine-tuning the model and top-$m$ as \{48 $\times$ 2, 96 $\times$ 2, 128 $\times$ 2, 192 $\times$ 2\} for a fixed cache size.
The inference performance and the corresponding training loss are presented in Table~\ref{tab:Cache Sizes} and Figure~\ref{fig:ablation1}, respectively. 

Figure~\ref{fig:ablation1} shows that when \( K \) values are set to 96, 128, and 192, the differences in training loss are minimal. This indicates that retaining tokens with the highest top-$K$ attention weights is sufficient, and further increasing $K$ does not yield substantial improvements in model performance.
From Table~\ref{tab:Cache Sizes}, it can be observed that there is no significant improvement in the quality of the generated text when \( m \) increases from \( 96 \times 2 \) to \( 192 \times 2 \), which, however, is accompanied by a notable decrease in throughput. 
Therefore, it is essential to select an appropriate set of $K$ and $m$, which balances throughput and the quality of generated text. 

\begin{table}[htbp]
\scriptsize
\centering
\setlength{\tabcolsep}{3.2mm} 
\begin{tabular}{*{6}{c|c|ccc}}
  \bottomrule
  $m$ & Throughput &BLEU&ROUGE&BERT-F \\
  \hline\hline 
  48 $\times$ 2 & \textbf{270.5} &35.5 & 27.8 & 62.8 \\
  96 $\times$ 2 & 259.6 &36.8 & 28.8 & 63.5 \\
  128 $\times$ 2 & 202.6 &37.0 & 29.2 & 63.7 \\
  192 $\times$ 2 & 167.7 &\textbf{37.3} & \textbf{29.4} & \textbf{64.3} \\
  \toprule
\end{tabular}
\vspace{-3mm}
\caption{Inference performance comparison of maintaining different cache sizes $m$ by adaptive token release. The best results are marked \textbf{bold}.}\label{tab:Cache Sizes}
\end{table}

\subsubsection{Influence of KV States Rebuild}
\begin{table}[htbp]
\tabcolsep 0.03in
\centering
\scriptsize
\setlength{\tabcolsep}{3.2mm} 
\begin{tabular}{*{6}{l|c|ccc}}
  \bottomrule
  Numbers & Throughput & BLEU&ROUGE&BERT-F \\
  \hline \hline
  $R$=0 & \textbf{278.2} &34.3 & 26.8 & 62.3 \\
  $R$=8 & 259.6 &36.8 & 28.8 & 63.5 \\
  $R$=16 & 202.6 &37.5 & 28.9 & 63.9 \\
  $R$=32 & 150.8 &\textbf{38.0} & \textbf{29.9} & \textbf{64.3} \\
  \toprule
\end{tabular}
\vspace{-3mm}
\caption{Inference performance comparison of different numbers of rebuilt tokens during inference. The best results are marked \textbf{bold}.}\label{tab:Re-computed}
\end{table}

We evaluate the impact of different $R$ in KV states rebuild. Specifically, we select $R$ in the range of \{0, 8, 16, 32\} and summarize the inference performance in Table~\ref{tab:Re-computed}. The results demonstrate that as the $R$ in rebuilt tokens increases, the model's performance first improves. However, the improvement comes at the cost of a reduction in throughput. When the number of rebuilt tokens is further increased from 16 to 32, we can observe an improvement of 1.5\% in BLEU, 1.0\% in ROUGH, and 0.4\% in BERT-F. 
However, this minor improvement is accompanied by a 34.4\% decrease in throughput. 
This indicates that selecting the appropriate number of rebuilding tokens is crucial for maintaining a trade-off between performance and quality during the inference process.

\subsubsection{Effectiveness of Controller Module} 
Since we use the controller module for advancedly predicting top-$K$ attention weights, next we investigate how it affects overall performance. In particular, we adjust the module with the following variants: 
\textbf{1)} w/o GRU: directly using the MLP for predicting the keeping/dropping probability of tokens; 
\textbf{2)} \name$_{d'=64}$: set hidden size of the controller to 64; 
\textbf{3)} \name$_{d'=128}$: set hidden size of the controller to 128; 
We first report accuracy and F1 scores on the dataset that fine-tunes the controller module, as detailed in Section~\ref{Adaptive Token Release with Controller Module}. Then, we report BLEU, ROUGE, and BERT-F scores on the Ultrachat benchmark, which further illustrate how the performance of the controller module influences the performance of LLMs.

We summarize the results in Table~\ref{tab:Controller}. Our observations are as follows: \textbf{1)} The GRU is crucial for the controller module to serve as an effective alternative to full attention; 
\textbf{2)} An improved controller module results in enhanced performance during the inference process, as it offers a more accurate approximation of sparse attention.

\begin{table}[htbp]
\tabcolsep 0.04in
\centering
\scriptsize
\setlength{\tabcolsep}{2.3mm} 
\begin{tabular}{l|cc|ccc}
  \bottomrule
 & \multicolumn{2}{c|}{Controller} & \multicolumn{3}{c}{Inference}\\
Variants &Acc.&F1&BLEU&ROUGE&BERT-F\\
  \hline\hline
    w/o GRU & 83.4 & 78.8 & 36.2 & 26.4 & 61.7 \\
\name$_{d'=128}$  & \textbf{87.9} & \textbf{82.3} & \textbf{37.5} & \textbf{28.9} & \textbf{63.9} \\
  \name$_{d'=64}$ & 81.5 & 74.0 & 33.5 & 28.5 & 62.4 \\
  \toprule
\end{tabular}
\vspace{-3mm}
\caption{Performance comparison of different variants controller module and inference. The best results are marked \textbf{bold}.}\label{tab:Controller}
\end{table}

\section{Related Work}
\label{sec:related_work}
\subsection{Sparse Attention}
Several works have attempted to integrate sparse attention into transformer-based models. This integration reduces the quadratic computational complexity in the sequence length, making it possible to process longer sequences. Some studies adopt fixed-pattern sparse strategies~\cite{zaheer2020big, beltagy2020longformer}, while others focus on sparsification based on the distribution and features of self-attention~\cite{goyal2020power, liu2023deja, huang2024sparse}. 
However, these methods either optimize bi-directional attention encoding, as in BERT, or fail to yield a practical improvement in the inference speed of language models~\cite{ren-etal-2023-context, tan2024sparsity}. This is because the reduction in the number of tokens does not yield significant benefits on CUDA~\cite{bolya2022token}. To address this issue, in the LLM inference process, we propose applying dynamic sparse attention to the storage of the KV cache, thereby fundamentally enhancing the throughput of the LLM.

\subsection{Efficient Inference for LLMs}
The efficiency improvement of LLM inference is becoming increasingly attention-grabbing~\cite{huang-chang-2023-towards}. 
Recent research has primarily focused on two aspects: systems and algorithms, aiming to enhance LLM inference efficiency. 
In recent years, numerous systems dedicated to LLM inference have emerged, such as FasterTransformer, Hugging Face Accelerate~\cite{accelerate}, FlexGen~\cite{sheng2023flexgen}, and vLLM~\cite{kwon2023efficient}. 
These systems often emphasize optimization from hardware accelerators and CUDA kernels. 
On the other hand, algorithms like Early-Exit~\cite{sun2021early, rotem-etal-2023-finding} Flashattention-2~\cite{dao2023flashattention} and Continuous Batch~\cite{yu2022orca} attempt to optimize LLM inference performance by reducing computational costs. 
In this paper, our proposed method is orthogonal to all mainstream LLM inference systems and most algorithmic optimizations, and our method can be used in parallel with these methods.

\subsection{Length Extrapolation for LLM Inference}
Length extrapolation aims to enable language models to maintain satisfactory performance when applied to super-long sequences as well.
Current research primarily focuses on finding improved representations for positional encoding. Rotary Position Embeddings (RoPE)~\cite{su2024roformer} attempt to transform absolute positions into relative position encodings for length expansion. Furthermore, ALiBi~\cite{press2021train} introduces relative positional information by imposing a penalty bias proportional to the distance in relative proximity on the attention matrix. However, current pproaches still struggle to model extremely long texts effectively. Simultaneously, when dealing with long texts, a major limiting factor lies in GPU memory overflow. In this paper, our approach extends the inference length of LLM by setting a fixed attention window size by adaptively releasing tokens, which is designed to maximize the length of inference without compromising performance significantly.

\section{Conclusion}
\label{sec:conclusion}
We propose an efficient sparse attention for the inference process of LLMs. This is achieved by adaptively releasing the KV state of the tokens with the lowest attention contribution in the cache while simultaneously rebuilding the state of tokens with the highest contribution during the step-by-step decoding of each token. Experimental results show that our approach significantly enhances the throughput of model inference without substantially compromising the quality of the generated text.

\section{Limitations}
In this paper, the primary limitation lies in the fine-tuning process required to align with our designed inference optimization method.
Specifically, during fine-tuning, we still face an $O(n^2)$ time complexity for self-attention, resulting in no speed improvement when learning dynamic sparse attention. 
Furthermore, our method is not immediately applicable during inference; it requires additional computational overhead for fine-tuning and training the controller to attain enhanced performance during inference.

\paragraph{Acknowledgments}
We are grateful for the funding support from the National Natural Science Foundation of China under Grant Numbers 62302345 and U23A20305, the Natural Science Foundation of Hubei Province under Grant Numbers 2023AFB192 and 2023BAB160, the Xiaomi Young Scholar Program, and the Wuhan University Talent Startup Fund.
\bibliography{ref}

\begin{thebibliography}{52}
\providecommand{\natexlab}[1]{#1}

\bibitem[{Aminabadi et~al.(2022)Aminabadi, Rajbhandari, Awan, Li, Li, Zheng, Ruwase, Smith, Zhang, Rasley et~al.}]{aminabadi2022deepspeed}
Reza~Yazdani Aminabadi, Samyam Rajbhandari, Ammar~Ahmad Awan, Cheng Li, Du~Li, Elton Zheng, Olatunji Ruwase, Shaden Smith, Minjia Zhang, Jeff Rasley, et~al. 2022.
\newblock Deepspeed-inference: enabling efficient inference of transformer models at unprecedented scale.
\newblock In \emph{SC22: International Conference for High Performance Computing, Networking, Storage and Analysis}, pages 1--15. IEEE.

\bibitem[{Beltagy et~al.(2020)Beltagy, Peters, and Cohan}]{beltagy2020longformer}
Iz~Beltagy, Matthew~E Peters, and Arman Cohan. 2020.
\newblock Longformer: The long-document transformer.
\newblock \emph{arXiv preprint arXiv:2004.05150}.

\bibitem[{Black et~al.(2021)Black, Gao, Wang, Leahy, and Biderman}]{gpt-neo}
Sid Black, Leo Gao, Phil Wang, Connor Leahy, and Stella Biderman. 2021.
\newblock \href {https://doi.org/10.5281/zenodo.5297715} {{GPT-Neo: Large Scale Autoregressive Language Modeling with Mesh-Tensorflow}}.
\newblock {If you use this software, please cite it using these metadata.}

\bibitem[{Bolya et~al.(2022)Bolya, Fu, Dai, Zhang, Feichtenhofer, and Hoffman}]{bolya2022token}
Daniel Bolya, Cheng-Yang Fu, Xiaoliang Dai, Peizhao Zhang, Christoph Feichtenhofer, and Judy Hoffman. 2022.
\newblock Token merging: Your vit but faster.
\newblock \emph{arXiv preprint arXiv:2210.09461}.

\bibitem[{Borzunov et~al.(2022)Borzunov, Baranchuk, Dettmers, Ryabinin, Belkada, Chumachenko, Samygin, and Raffel}]{borzunov2022petals}
Alexander Borzunov, Dmitry Baranchuk, Tim Dettmers, Max Ryabinin, Younes Belkada, Artem Chumachenko, Pavel Samygin, and Colin Raffel. 2022.
\newblock Petals: Collaborative inference and fine-tuning of large models.
\newblock \emph{arXiv preprint arXiv:2209.01188}.

\bibitem[{Child et~al.(2019)Child, Gray, Radford, and Sutskever}]{child2019sparsetransformer}
Rewon Child, Scott Gray, Alec Radford, and Ilya Sutskever. 2019.
\newblock Generating long sequences with sparse transformers.
\newblock \emph{URL https://openai.com/blog/sparse-transformers}.

\bibitem[{Dao(2023)}]{dao2023flashattention}
Tri Dao. 2023.
\newblock Flashattention-2: Faster attention with better parallelism and work partitioning.
\newblock \emph{arXiv preprint arXiv:2307.08691}.

\bibitem[{Dettmers et~al.(2023)Dettmers, Pagnoni, Holtzman, and Zettlemoyer}]{dettmers2023qlora}
Tim Dettmers, Artidoro Pagnoni, Ari Holtzman, and Luke Zettlemoyer. 2023.
\newblock Qlora: Efficient finetuning of quantized llms.
\newblock \emph{arXiv preprint arXiv:2305.14314}.

\bibitem[{Dey and Salem(2017)}]{dey2017gate}
Rahul Dey and Fathi~M Salem. 2017.
\newblock Gate-variants of gated recurrent unit (gru) neural networks.
\newblock In \emph{2017 IEEE 60th international midwest symposium on circuits and systems (MWSCAS)}, pages 1597--1600. IEEE.

\bibitem[{Ding et~al.(2023)Ding, Chen, Xu, Qin, Zheng, Hu, Liu, Sun, and Zhou}]{ding2023enhancing}
Ning Ding, Yulin Chen, Bokai Xu, Yujia Qin, Zhi Zheng, Shengding Hu, Zhiyuan Liu, Maosong Sun, and Bowen Zhou. 2023.
\newblock Enhancing chat language models by scaling high-quality instructional conversations.
\newblock \emph{arXiv preprint arXiv:2305.14233}.

\bibitem[{Gliwa et~al.(2019)Gliwa, Mochol, Biesek, and Wawer}]{gliwa-etal-2019-samsum}
Bogdan Gliwa, Iwona Mochol, Maciej Biesek, and Aleksander Wawer. 2019.
\newblock \href {https://doi.org/10.18653/v1/D19-5409} {{SAMS}um corpus: A human-annotated dialogue dataset for abstractive summarization}.
\newblock In \emph{Proceedings of the 2nd Workshop on New Frontiers in Summarization}, pages 70--79, Hong Kong, China. Association for Computational Linguistics.

\bibitem[{Goyal et~al.(2020)Goyal, Choudhury, Raje, Chakaravarthy, Sabharwal, and Verma}]{goyal2020power}
Saurabh Goyal, Anamitra~Roy Choudhury, Saurabh Raje, Venkatesan Chakaravarthy, Yogish Sabharwal, and Ashish Verma. 2020.
\newblock Power-bert: Accelerating bert inference via progressive word-vector elimination.
\newblock In \emph{International Conference on Machine Learning}, pages 3690--3699. PMLR.

\bibitem[{Gugger et~al.(2022)Gugger, Debut, Wolf, Schmid, Mueller, Mangrulkar, Sun, and Bossan}]{accelerate}
Sylvain Gugger, Lysandre Debut, Thomas Wolf, Philipp Schmid, Zachary Mueller, Sourab Mangrulkar, Marc Sun, and Benjamin Bossan. 2022.
\newblock Accelerate: Training and inference at scale made simple, efficient and adaptable.
\newblock \url{https://github.com/huggingface/accelerate}.

\bibitem[{Hassani et~al.(2023)Hassani, Walton, Li, Li, and Shi}]{hassani2023neighborhood}
Ali Hassani, Steven Walton, Jiachen Li, Shen Li, and Humphrey Shi. 2023.
\newblock Neighborhood attention transformer.
\newblock In \emph{Proceedings of the IEEE/CVF Conference on Computer Vision and Pattern Recognition}, pages 6185--6194.

\bibitem[{Huang and Chang(2023)}]{huang-chang-2023-towards}
Jie Huang and Kevin Chen-Chuan Chang. 2023.
\newblock \href {https://doi.org/10.18653/v1/2023.findings-acl.67} {Towards reasoning in large language models: A survey}.
\newblock In \emph{Findings of the Association for Computational Linguistics: ACL 2023}, pages 1049--1065, Toronto, Canada. Association for Computational Linguistics.

\bibitem[{Huang et~al.(2024)Huang, Deng, Hui, Wu, Zhou, and Wang}]{huang2024sparse}
Wenli Huang, Ye~Deng, Siqi Hui, Yang Wu, Sanping Zhou, and Jinjun Wang. 2024.
\newblock Sparse self-attention transformer for image inpainting.
\newblock \emph{Pattern Recognition}, 145:109897.

\bibitem[{Kaplan et~al.(2020)Kaplan, McCandlish, Henighan, Brown, Chess, Child, Gray, Radford, Wu, and Amodei}]{kaplan2020scaling}
Jared Kaplan, Sam McCandlish, Tom Henighan, Tom~B Brown, Benjamin Chess, Rewon Child, Scott Gray, Alec Radford, Jeffrey Wu, and Dario Amodei. 2020.
\newblock Scaling laws for neural language models.
\newblock \emph{arXiv preprint arXiv:2001.08361}.

\bibitem[{Kim et~al.(2022)Kim, Wimmer, and Kim}]{kim2022analysis}
Seongsoo Kim, Hayden Wimmer, and Jongyeop Kim. 2022.
\newblock Analysis of deep learning libraries: Keras, pytorch, and mxnet.
\newblock In \emph{2022 IEEE/ACIS 20th International Conference on Software Engineering Research, Management and Applications (SERA)}, pages 54--62. IEEE.

\bibitem[{Kwon et~al.(2023)Kwon, Li, Zhuang, Sheng, Zheng, Yu, Gonzalez, Zhang, and Stoica}]{kwon2023efficient}
Woosuk Kwon, Zhuohan Li, Siyuan Zhuang, Ying Sheng, Lianmin Zheng, Cody~Hao Yu, Joseph~E. Gonzalez, Hao Zhang, and Ion Stoica. 2023.
\newblock Efficient memory management for large language model serving with pagedattention.
\newblock In \emph{Proceedings of the ACM SIGOPS 29th Symposium on Operating Systems Principles}.

\bibitem[{Li et~al.(2023{\natexlab{a}})Li, Wang, Liu, Zhao, Wang, Wang, Zou, Fan, and Li}]{10.1145/3604915.3608779}
Chengxi Li, Yejing Wang, Qidong Liu, Xiangyu Zhao, Wanyu Wang, Yiqi Wang, Lixin Zou, Wenqi Fan, and Qing Li. 2023{\natexlab{a}}.
\newblock \href {https://doi.org/10.1145/3604915.3608779} {Strec: Sparse transformer for sequential recommendations}.
\newblock In \emph{Proceedings of the 17th ACM Conference on Recommender Systems}, RecSys '23, page 101–111, New York, NY, USA. Association for Computing Machinery.

\bibitem[{Li et~al.(2023{\natexlab{b}})Li, Shao, and Xie}]{li2023long1}
Dacheng Li, Rulin Shao, and Anze~and Xie. 2023{\natexlab{b}}.
\newblock How long can context length of open-source llms truly promise?
\newblock In \emph{NeurIPS 2023 Workshop on Instruction Tuning and Instruction Following}.

\bibitem[{Li et~al.(2023{\natexlab{c}})Li, Hammoud, Itani, Khizbullin, and Ghanem}]{li2023camel}
Guohao Li, Hasan Abed Al~Kader Hammoud, Hani Itani, Dmitrii Khizbullin, and Bernard Ghanem. 2023{\natexlab{c}}.
\newblock Camel: Communicative agents for" mind" exploration of large scale language model society.
\newblock \emph{arXiv preprint arXiv:2303.17760}.

\bibitem[{Li et~al.(2023{\natexlab{d}})Li, Keung, Cheng, Kasai, and Smith}]{li-etal-2023-narrowbert}
Haoxin Li, Phillip Keung, Daniel Cheng, Jungo Kasai, and Noah~A. Smith. 2023{\natexlab{d}}.
\newblock \href {https://doi.org/10.18653/v1/2023.acl-short.146} {{N}arrow{BERT}: Accelerating masked language model pretraining and inference}.
\newblock In \emph{Proceedings of the 61st Annual Meeting of the Association for Computational Linguistics (Volume 2: Short Papers)}, pages 1723--1730, Toronto, Canada. Association for Computational Linguistics.

\bibitem[{Li et~al.(2023{\natexlab{e}})Li, Dong, Guerin, and Lin}]{li-etal-2023-compressing}
Yucheng Li, Bo~Dong, Frank Guerin, and Chenghua Lin. 2023{\natexlab{e}}.
\newblock \href {https://doi.org/10.18653/v1/2023.emnlp-main.391} {Compressing context to enhance inference efficiency of large language models}.
\newblock In \emph{Proceedings of the 2023 Conference on Empirical Methods in Natural Language Processing}, pages 6342--6353, Singapore. Association for Computational Linguistics.

\bibitem[{Liang et~al.(2023)Liang, Li, Wu, Cao, and Zhang}]{liang-etal-2023-dynamic}
Xiaobo Liang, Juntao Li, Lijun Wu, Ziqiang Cao, and Min Zhang. 2023.
\newblock \href {https://doi.org/10.18653/v1/2023.acl-long.162} {Dynamic and efficient inference for text generation via {BERT} family}.
\newblock In \emph{Proceedings of the 61st Annual Meeting of the Association for Computational Linguistics (Volume 1: Long Papers)}, pages 2883--2897, Toronto, Canada. Association for Computational Linguistics.

\bibitem[{Liu et~al.(2022)Liu, Ye, Ren, and Wang}]{liu2022dynast}
Songhua Liu, Jingwen Ye, Sucheng Ren, and Xinchao Wang. 2022.
\newblock Dynast: Dynamic sparse transformer for exemplar-guided image generation.
\newblock In \emph{European Conference on Computer Vision}, pages 72--90. Springer.

\bibitem[{Liu et~al.(2023{\natexlab{a}})Liu, Desai, Liao, Wang, Xie, Xu, Kyrillidis, and Shrivastava}]{liu2023scissorhands}
Zichang Liu, Aditya Desai, Fangshuo Liao, Weitao Wang, Victor Xie, Zhaozhuo Xu, Anastasios Kyrillidis, and Anshumali Shrivastava. 2023{\natexlab{a}}.
\newblock Scissorhands: Exploiting the persistence of importance hypothesis for llm kv cache compression at test time.
\newblock \emph{arXiv preprint arXiv:2305.17118}.

\bibitem[{Liu et~al.(2023{\natexlab{b}})Liu, Wang, Dao, Zhou, Yuan, Song, Shrivastava, Zhang, Tian, Re et~al.}]{liu2023deja}
Zichang Liu, Jue Wang, Tri Dao, Tianyi Zhou, Binhang Yuan, Zhao Song, Anshumali Shrivastava, Ce~Zhang, Yuandong Tian, Christopher Re, et~al. 2023{\natexlab{b}}.
\newblock Deja vu: Contextual sparsity for efficient llms at inference time.
\newblock In \emph{International Conference on Machine Learning}, pages 22137--22176. PMLR.

\bibitem[{Mao et~al.(2024)Mao, Zou, Zheng, Tang, Chu, Zhao, Wang, and Yin}]{10.1145/3589334.3645474}
Haitao Mao, Lixin Zou, Yujia Zheng, Jiliang Tang, Xiaokai Chu, Jiashu Zhao, Qian Wang, and Dawei Yin. 2024.
\newblock \href {https://doi.org/10.1145/3589334.3645474} {Whole page unbiased learning to rank}.
\newblock In \emph{Proceedings of the ACM on Web Conference 2024}, WWW '24, page 1431–1440, New York, NY, USA. Association for Computing Machinery.

\bibitem[{Pan et~al.(2024)Pan, Wu, Jiang, Xia, Luo, Zhang, Lin, R{\"u}hle, Yang, Lin et~al.}]{pan2024llmlingua}
Zhuoshi Pan, Qianhui Wu, Huiqiang Jiang, Menglin Xia, Xufang Luo, Jue Zhang, Qingwei Lin, Victor R{\"u}hle, Yuqing Yang, Chin-Yew Lin, et~al. 2024.
\newblock Llmlingua-2: Data distillation for efficient and faithful task-agnostic prompt compression.
\newblock \emph{arXiv preprint arXiv:2403.12968}.

\bibitem[{Press et~al.(2021)Press, Smith, and Lewis}]{press2021train}
Ofir Press, Noah~A Smith, and Mike Lewis. 2021.
\newblock Train short, test long: Attention with linear biases enables input length extrapolation.
\newblock \emph{arXiv preprint arXiv:2108.12409}.

\bibitem[{Rao et~al.(2021)Rao, Zhao, Liu, Lu, Zhou, and Hsieh}]{rao2021dynamicvit}
Yongming Rao, Wenliang Zhao, Benlin Liu, Jiwen Lu, Jie Zhou, and Cho-Jui Hsieh. 2021.
\newblock Dynamicvit: Efficient vision transformers with dynamic token sparsification.
\newblock \emph{Advances in neural information processing systems}, 34:13937--13949.

\bibitem[{Rasley et~al.(2020)Rasley, Rajbhandari, Ruwase, and He}]{rasley2020deepspeed}
Jeff Rasley, Samyam Rajbhandari, Olatunji Ruwase, and Yuxiong He. 2020.
\newblock Deepspeed: System optimizations enable training deep learning models with over 100 billion parameters.
\newblock In \emph{Proceedings of the 26th ACM SIGKDD International Conference on Knowledge Discovery \& Data Mining}, pages 3505--3506.

\bibitem[{Ren et~al.(2023)Ren, Jia, and Zhu}]{ren-etal-2023-context}
Siyu Ren, Qi~Jia, and Kenny Zhu. 2023.
\newblock \href {https://doi.org/10.18653/v1/2023.emnlp-main.794} {Context compression for auto-regressive transformers with sentinel tokens}.
\newblock In \emph{Proceedings of the 2023 Conference on Empirical Methods in Natural Language Processing}, pages 12860--12867, Singapore. Association for Computational Linguistics.

\bibitem[{Rotem et~al.(2023)Rotem, Hassid, Mamou, and Schwartz}]{rotem-etal-2023-finding}
Daniel Rotem, Michael Hassid, Jonathan Mamou, and Roy Schwartz. 2023.
\newblock \href {https://doi.org/10.18653/v1/2023.acl-long.829} {Finding the {SWEET} spot: Analysis and improvement of adaptive inference in low resource settings}.
\newblock In \emph{Proceedings of the 61st Annual Meeting of the Association for Computational Linguistics (Volume 1: Long Papers)}, pages 14836--14851, Toronto, Canada. Association for Computational Linguistics.

\bibitem[{See et~al.(2017)See, Liu, and Manning}]{see-etal-2017-get}
Abigail See, Peter~J. Liu, and Christopher~D. Manning. 2017.
\newblock \href {https://doi.org/10.18653/v1/P17-1099} {Get to the point: Summarization with pointer-generator networks}.
\newblock In \emph{Proceedings of the 55th Annual Meeting of the Association for Computational Linguistics (Volume 1: Long Papers)}, pages 1073--1083, Vancouver, Canada. Association for Computational Linguistics.

\bibitem[{Sheng et~al.(2023)Sheng, Zheng, Yuan, Li, Ryabinin, Chen, Liang, R{\'e}, Stoica, and Zhang}]{sheng2023flexgen}
Ying Sheng, Lianmin Zheng, Binhang Yuan, Zhuohan Li, Max Ryabinin, Beidi Chen, Percy Liang, Christopher R{\'e}, Ion Stoica, and Ce~Zhang. 2023.
\newblock Flexgen: High-throughput generative inference of large language models with a single gpu.
\newblock In \emph{International Conference on Machine Learning}, pages 31094--31116. PMLR.

\bibitem[{Strati et~al.(2024)Strati, Mcallister, Phanishayee, Tarnawski, and Klimovic}]{strati2024d}
Foteini Strati, Sara Mcallister, Amar Phanishayee, Jakub Tarnawski, and Ana Klimovic. 2024.
\newblock D$\backslash$'ej$\backslash$avu: Kv-cache streaming for fast, fault-tolerant generative llm serving.
\newblock \emph{arXiv preprint arXiv:2403.01876}.

\bibitem[{Su et~al.(2024)Su, Ahmed, Lu, Pan, Bo, and Liu}]{su2024roformer}
Jianlin Su, Murtadha Ahmed, Yu~Lu, Shengfeng Pan, Wen Bo, and Yunfeng Liu. 2024.
\newblock Roformer: Enhanced transformer with rotary position embedding.
\newblock \emph{Neurocomputing}, 568:127063.

\bibitem[{Sun et~al.(2021)Sun, Zhou, Liu, Zhang, Jiang, Cao, Huang, and Qiu}]{sun2021early}
Tianxiang Sun, Yunhua Zhou, Xiangyang Liu, Xinyu Zhang, Hao Jiang, Zhao Cao, Xuanjing Huang, and Xipeng Qiu. 2021.
\newblock Early exiting with ensemble internal classifiers.
\newblock \emph{arXiv preprint arXiv:2105.13792}.

\bibitem[{Tan et~al.(2024)Tan, Chen, Zhang, and Liu}]{tan2024sparsity}
Zhen Tan, Tianlong Chen, Zhenyu Zhang, and Huan Liu. 2024.
\newblock Sparsity-guided holistic explanation for llms with interpretable inference-time intervention.
\newblock In \emph{Proceedings of the AAAI Conference on Artificial Intelligence}, volume~38, pages 21619--21627.

\bibitem[{Tang et~al.(2023)Tang, Wang, Zou, Zhang, Zhou, and Li}]{10.1145/3539618.3591778}
Zuoli Tang, Lin Wang, Lixin Zou, Xiaolu Zhang, Jun Zhou, and Chenliang Li. 2023.
\newblock \href {https://doi.org/10.1145/3539618.3591778} {Towards multi-interest pre-training with sparse capsule network}.
\newblock In \emph{Proceedings of the 46th International ACM SIGIR Conference on Research and Development in Information Retrieval}, SIGIR '23, page 311–320, New York, NY, USA. Association for Computing Machinery.

\bibitem[{Touvron et~al.(2023)Touvron, Martin, Stone, Albert, Almahairi, Babaei, Bashlykov, Batra, Bhargava, Bhosale et~al.}]{touvron2023llama}
Hugo Touvron, Louis Martin, Kevin Stone, Peter Albert, Amjad Almahairi, Yasmine Babaei, Nikolay Bashlykov, Soumya Batra, Prajjwal Bhargava, Shruti Bhosale, et~al. 2023.
\newblock Llama 2: Open foundation and fine-tuned chat models.
\newblock \emph{arXiv preprint arXiv:2307.09288}.

\bibitem[{Vu et~al.(2024)Vu, Wang, Li, Chen, Zhao, Xing, and Chen}]{vu2024gptvoicetasker}
Minh~Duc Vu, Han Wang, Zhuang Li, Jieshan Chen, Shengdong Zhao, Zhenchang Xing, and Chunyang Chen. 2024.
\newblock Gptvoicetasker: Llm-powered virtual assistant for smartphone.
\newblock \emph{arXiv preprint arXiv:2401.14268}.

\bibitem[{Wu et~al.(2023)Wu, Zeng, Wang, Wang, and Chen}]{wu2023ppt}
Xinjian Wu, Fanhu Zeng, Xiudong Wang, Yunhe Wang, and Xinghao Chen. 2023.
\newblock Ppt: Token pruning and pooling for efficient vision transformers.
\newblock \emph{arXiv preprint arXiv:2310.01812}.

\bibitem[{Xiao et~al.(2024)Xiao, Tian, Chen, Han, and Lewis}]{xiao2024efficient}
Guangxuan Xiao, Yuandong Tian, Beidi Chen, Song Han, and Mike Lewis. 2024.
\newblock \href {https://openreview.net/forum?id=NG7sS51zVF} {Efficient streaming language models with attention sinks}.
\newblock In \emph{The Twelfth International Conference on Learning Representations}.

\bibitem[{Yang et~al.(2023)Yang, Lee, Cho, Papailiopoulos, and Lee}]{yang2023predictive}
Seongjun Yang, Gibbeum Lee, Jaewoong Cho, Dimitris Papailiopoulos, and Kangwook Lee. 2023.
\newblock Predictive pipelined decoding: A compute-latency trade-off for exact llm decoding.
\newblock \emph{arXiv preprint arXiv:2307.05908}.

\bibitem[{Yu et~al.(2022)Yu, Jeong, Kim, Kim, and Chun}]{yu2022orca}
Gyeong-In Yu, Joo~Seong Jeong, Geon-Woo Kim, Soojeong Kim, and Byung-Gon Chun. 2022.
\newblock Orca: A distributed serving system for $\{$Transformer-Based$\}$ generative models.
\newblock In \emph{16th USENIX Symposium on Operating Systems Design and Implementation (OSDI 22)}, pages 521--538.

\bibitem[{Zaheer et~al.(2020)Zaheer, Guruganesh, Dubey, Ainslie, Alberti, Ontanon, Pham, Ravula, Wang, Yang et~al.}]{zaheer2020big}
Manzil Zaheer, Guru Guruganesh, Kumar~Avinava Dubey, Joshua Ainslie, Chris Alberti, Santiago Ontanon, Philip Pham, Anirudh Ravula, Qifan Wang, Li~Yang, et~al. 2020.
\newblock Big bird: Transformers for longer sequences.
\newblock \emph{Advances in neural information processing systems}, 33:17283--17297.

\bibitem[{Zhang et~al.(2019)Zhang, Kishore, Wu, Weinberger, and Artzi}]{zhang2019bertscore}
Tianyi Zhang, Varsha Kishore, Felix Wu, Kilian~Q Weinberger, and Yoav Artzi. 2019.
\newblock Bertscore: Evaluating text generation with bert.
\newblock \emph{arXiv preprint arXiv:1904.09675}.

\bibitem[{Zhang et~al.(2024)Zhang, Sheng, Zhou, Chen, Zheng, Cai, Song, Tian, R{\'e}, Barrett et~al.}]{zhang2024h2o}
Zhenyu Zhang, Ying Sheng, Tianyi Zhou, Tianlong Chen, Lianmin Zheng, Ruisi Cai, Zhao Song, Yuandong Tian, Christopher R{\'e}, Clark Barrett, et~al. 2024.
\newblock H2o: Heavy-hitter oracle for efficient generative inference of large language models.
\newblock \emph{Advances in Neural Information Processing Systems}, 36.

\bibitem[{Zhao et~al.(2023)Zhao, Zou, Zhao, Wang, and Yin}]{10.1145/3543507.3583418}
Kesen Zhao, Lixin Zou, Xiangyu Zhao, Maolin Wang, and Dawei Yin. 2023.
\newblock \href {https://doi.org/10.1145/3543507.3583418} {User retention-oriented recommendation with decision transformer}.
\newblock In \emph{Proceedings of the ACM Web Conference 2023}, WWW '23, page 1141–1149, New York, NY, USA. Association for Computing Machinery.

\end{thebibliography}

\clearpage
\appendix
\section{Dataset Statistics}
In Table~\ref{tab:data statistic}, we present the statistical information of the datasets used in our experiments, including dataset partitioning and sequence length statistics.
\begin{table}[h]
\centering
\scriptsize
\setlength{\tabcolsep}{0.6mm} 
 \renewcommand{\arraystretch}{1} 
\begin{tabular}{llllllll}
  \toprule
  \multirow{2}*{Dataset} & \multicolumn{3}{c}{Dataset partitioning} & \multicolumn{3}{c}{Sequence Length}\\
  \cmidrule(lr){2-4}\cmidrule(lr){5-7}
  &Train&Valid&Test&Average&Median&90 percentile\\
  \midrule
UltraChat&696600&77400&77400&1476&1411&2265\\
EverythingLM&972&108&108&1743&1765&2550\\
Math&45000&5000&5000&510&459&910\\
StreamEval&2825&353&352&1686&1679&2160\\
CNN Daily Mail&287113&13368&11490&1132&1060&1825\\
SAMSum&14700&818&819&3227&3212&3900\\
  \bottomrule
\end{tabular}
\vspace{-3mm}
\caption{Statistics of datasets. Sequence length measured in tokens using a SentencePiece Model.}\label{tab:data statistic}
\end{table}
\label{sec:Dataset Statistics}

\section{Implementation Details}
\label{sec:implement-detail}
In this section, we illustrate the details of our implementation, primarily including training data collection for the controller module, fine-tuning with QLoRA, details of the controller module, inference settings, and hardware settings.

\paragraph{Training data collection for the controller module} As detailed in Section~\ref{Adaptive Token Release with Controller Module}, during the fine-tuning process of the full attention (baseline) on the training set, we collect the word embeddings and the most frequently top-$K$ indices of each sample as input data and labels for training the controller. Given that full attention models the entire sequence, it consistently yields the lowest loss during fine-tuning, thereby ensuring that the attention distribution modeled is reliable and informative for capturing the top-$K$ tokens. 

\paragraph{Fine-tuning with QLoRA}(\romannumeral1) Hyper-parameters: For all methods, we utilize the Adam optimizer with a learning rate of 3e-5, decayed by a rate of 0.98 every 40 steps. Regarding the parameters for Q-LORA, we uniformly set the rank parameter \( r = 16 \) and the learning rate scaling factor \( \text{lora\_alpha} = 32 \). (\romannumeral2) Alignment fine-tuning with \name: By collecting the top-$K$ indices, we create attention masks for full attention, which block the attention from the current token to the low-contribution tokens. This implementation achieves dynamic sparse attention during fine-tuning, resulting in a model aligned with our inference optimization approach. 
The fine-tuning time on the UltraChat, EverythingLM, and Math datasets is approximately 3 hours on four GeForce RTX 3090 GPUs, respectively.

\paragraph{Details of the controller module} (a) Controller network structure: (\romannumeral1) Input layer: A GRU layer with an input size of 4096 and a hidden size of 128; (\romannumeral2) Position layer: A fully connected layer with an input size of 1, projected to 128; (\romannumeral3) Interaction layer: A fully connected layer with a hidden size of 128 and a Tanh activation function; (\romannumeral4) Output layer: Each output, mapped to [0,1] for cross-entropy loss over the sequence length, is obtained through a fully connected layer followed by a sigmoid function.
(b) Training Details: We employ the Adam optimizer with a learning rate of 0.005, accompanied by a decay rate of 0.98 every 2000 steps. We split the collected dataset into a training set and a validation set with an 8:2 ratio, and save the model parameters achieving the highest F1 score on the validation set.

\paragraph{Hardware settings} We utilize four GeForce RTX 3090 GPUs, with a total runtime exceeding 20 hours.

\begin{figure}[h]
\centering
\subfigure[Proportion of overlapping tokens between the uniform token set and top-$K/2$ tokens sets at each layer.]{
\includegraphics[width=3.3cm]{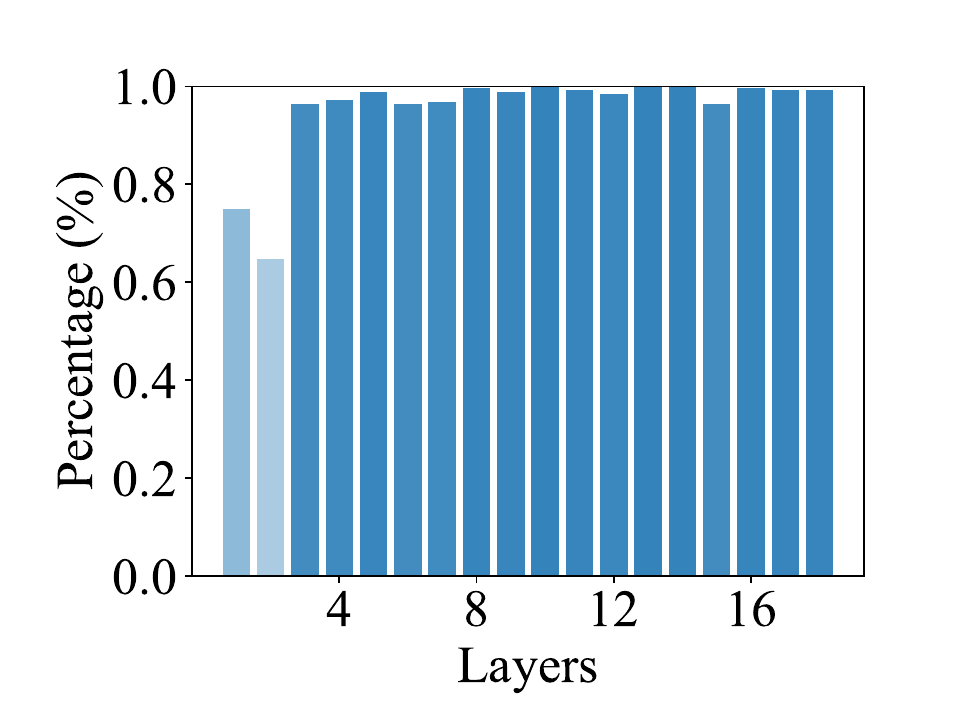}
\label{fig:attention_frequently1}
}
\hspace{1.5mm}
\subfigure[Proportion of cumulative softmax scores from the uniform token set at each layer.]{
\includegraphics[width=3.3cm]{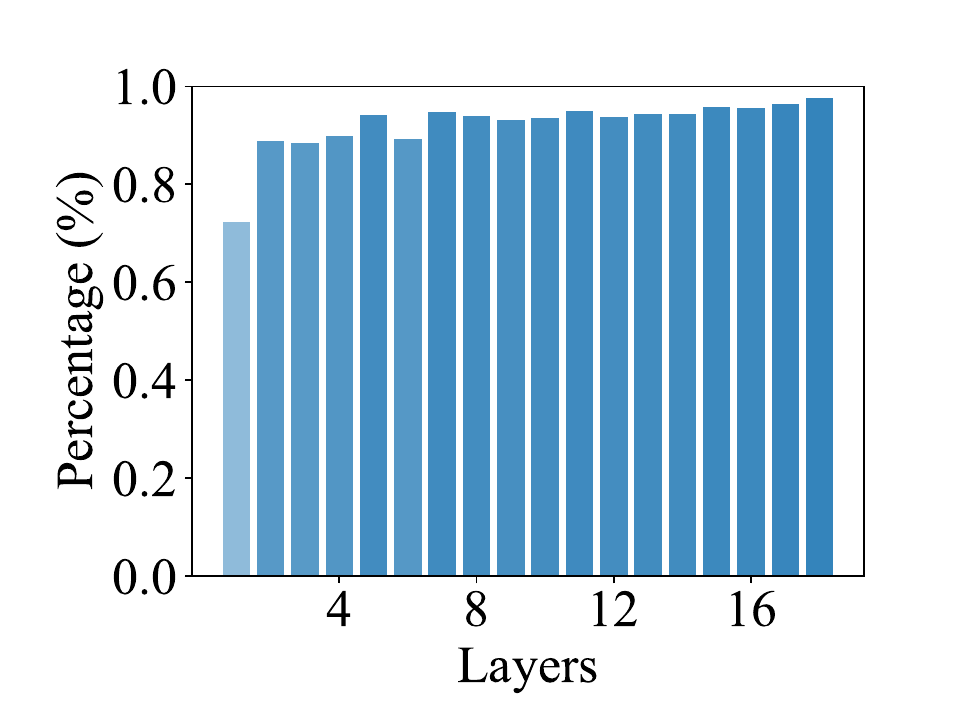}
\label{fig:attention_frequently2}
}
\caption{The effectiveness of the uniform token set at each layer.}
\label{fig:attention_frequently}
\end{figure}
\begin{figure*}[h]
\centerline{\includegraphics[width=1.0\textwidth]{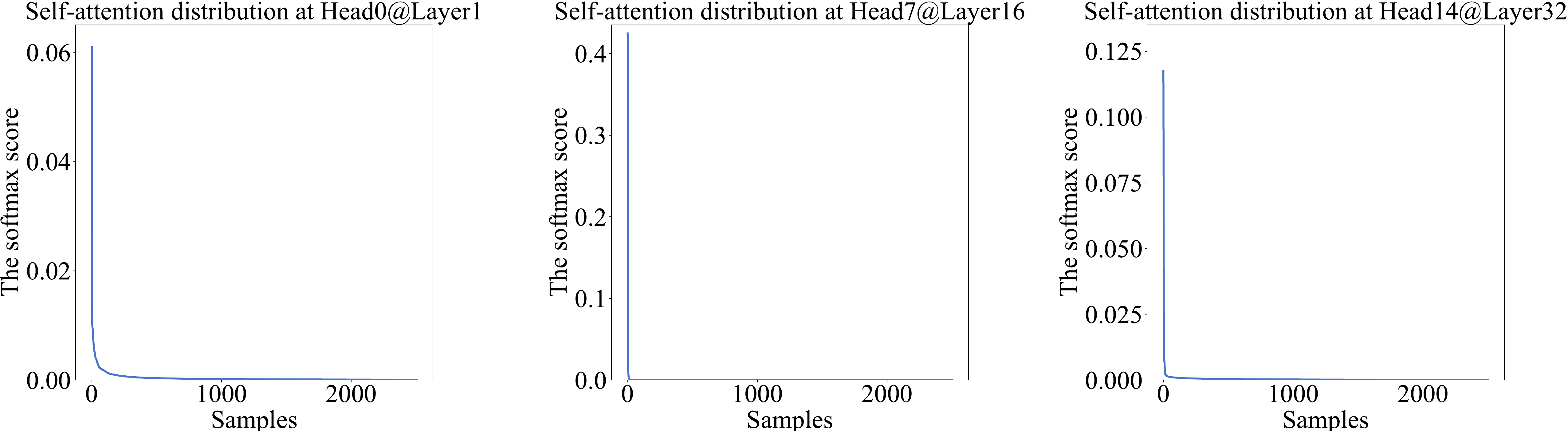}}
\caption{The Softmax scores in the self-attention from a
32-layer Transformer on CNN-DM dataset.}
\label{fig:attention_distribution}
\end{figure*}

\begin{figure*}[h]
\centerline{\includegraphics[width=1.0\textwidth]{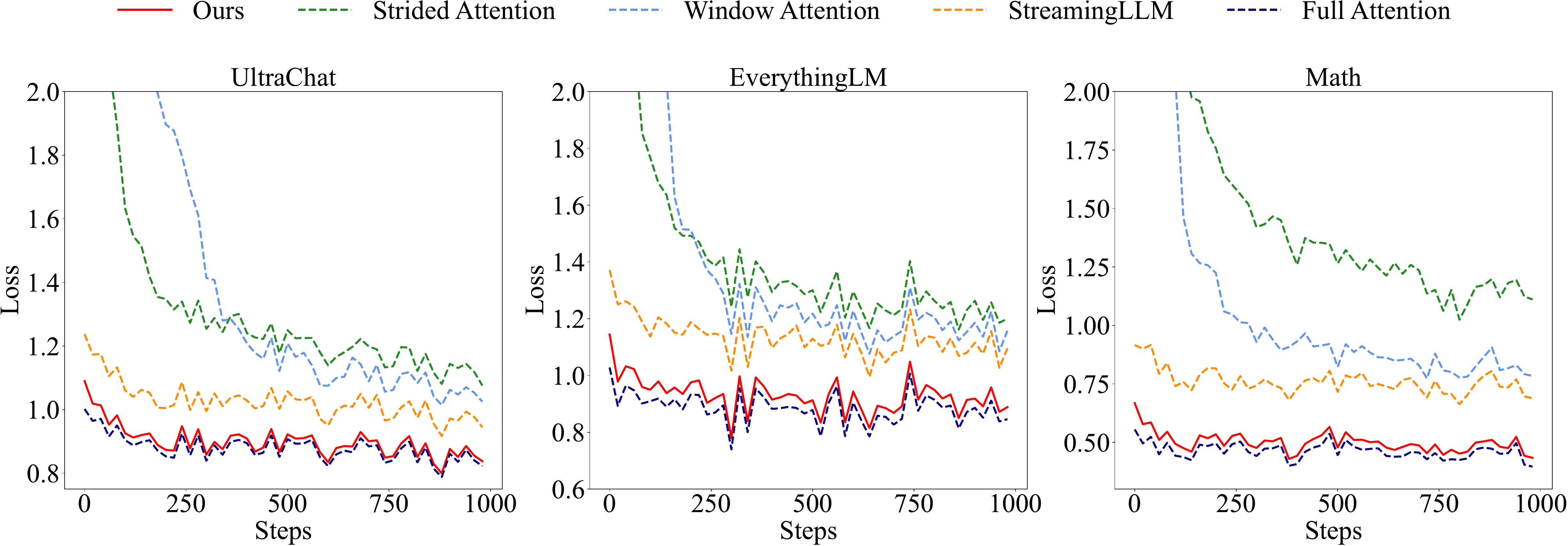}}
\caption{Comparison of loss during fine-tuning across different methods.}
\label{fig:exp1-loss}
\end{figure*}

\section{Analysis of Dynamic Sparse Attention}
\label{Analysis of Dynamic Sparse Attention}
In this section, we first demonstrate the effectiveness of applying a uniform scheduling policy across different layers. Then we showcase the superior performance of dynamic sparse attention in comparison to other methods and delve into the underlying reasons behind its effectiveness.

According to the settings in Section~\ref{Adaptive Token Release with Controller Module}, we select the top-$K/2$ tokens sets for each layer and consider the top-$K$ tokens that appear most frequently in these sets as the uniform token sets. In Figure~\ref{fig:attention_frequently1}, we observe that the uniform token set covers the majority of the top-K/2 token sets at each layer. Additionally, in Figure~\ref{fig:attention_frequently2}, we illustrate the cumulative softmax attention scores from the uniform token set for the current token across different layers, demonstrating that the uniform token set can effectively replace the contributions of the top-K/2 token sets at each layer.

Figure~\ref{fig:exp1-loss} illustrates the comparison of loss with QLoRA fine-tuning for various methods on UltraChat, EverythingLM, and Math. It is evident that the loss by focusing on the tokens with the top-$K$ highest attention (dynamic sparse attention) maintains consistency with the full attention approach and results in a notable reduction in loss compared to other methods.

The superior performance of dynamic sparse attention can be attributed to the observation that only a small portion of tokens significantly contributes to the attention mechanism during the modeling process for the current token. The Softmax scores in the self-attention curve on the cnn-daily dataset are presented in Figure~\ref{fig:attention_distribution}. It can be observed that the blue curve in Figure~\ref{fig:attention_distribution} forms a long-tail distribution, i.e. a few tokens contribute to the significant attention and others can be disregarded.

\section{Analysis of the unidirectional GRU}
\label{Comparison of unidirectional and bidirectional GRU performance}
In this section, we explore the impact of the GRU unit for adaptive token release and selection on runtime. Table~\ref{tab:time of GRU} reports the time needed to generate 100, 200, and 500 tokens, as well as the time allocated to the GRU unit for adaptive token release and selection. It can be
seen that the GRU unit's average runtime is merely 2.9\% of the total runtime. This suggests that the runtime overhead attributed to the GRU unit for adaptive token release and selection is negligible.
\begin{table}[h]
\centering
\setlength{\tabcolsep}{3mm} 
\begin{tabular}{*{4}{c|ccc}}
  \bottomrule
Generated Length & 100 & 200 & 500 \\
  \hline\hline
  runtime of GRU & 0.3 & 0.8 & 2.1 \\
  total runtime & 13.3 & 24.8 & 62.5 \\
  \toprule
\end{tabular}
\vspace{-3mm}
\caption{Runtime (seconds) of GRU and total inference process for generating different text lengths.}\label{tab:time of GRU}
\end{table}

In addition, we compare the performance of unidirectional and bidirectional GRU in terms of top-$K$ prediction Accuracy, F1-score, and Runtime (seconds) for 500 tokens to illustrate why we choose unidirectional GRU as the primary architecture for the controller module.

Table~\ref{tab:unidirectional and bidirectional GRU} demonstrates the Accuracy, F1-score, and Runtime. We can observe that bidirectional GRU does not significantly improve performance compared to unidirectional GRU. Instead, bidirectional GRU is more computationally expensive in terms of runtime because it requires forward and backward computations at each time step.

\begin{table}[h]
\centering
\setlength{\tabcolsep}{3mm} 
\begin{tabular}{*{4}{c|ccc}}
  \bottomrule
Model & Acc. & F1 & Runtime \\
  \hline\hline
  unidirectional GRU & 87.9 & 82.3 & 2.1 \\
  bidirectional GRU & 88.4 & 83.8 & 3.9 \\
  \toprule
\end{tabular}
\vspace{-3mm}
\caption{Performance comparison of unidirectional and bidirectional GRU}\label{tab:unidirectional and bidirectional GRU}
\end{table}

\end{document}